\pdfoutput=1
\documentclass[iicol, pdflatex, sn-basic, Numbered]{sn-jnl}

\usepackage{graphicx}%
\usepackage{multirow}%
\usepackage{amsmath,amssymb,amsfonts}%
\usepackage{amsthm}%
\usepackage{mathrsfs}%
\usepackage[title]{appendix}%
\usepackage{xcolor}%
\usepackage{textcomp}%
\usepackage{manyfoot}%
\usepackage{booktabs}%
\usepackage{algorithm}%
\usepackage{algorithmicx}%
\usepackage{algpseudocode}%
\usepackage{listings}%
\usepackage{enumitem}
\usepackage{times} 

\theoremstyle{thmstyleone}%
\theoremstyle{thmstyletwo}%

\theoremstyle{thmstylethree}%

\begin{document}

\title[Article Title]{DynamicDTA: Drug-Target Binding Affinity Prediction Using Dynamic Descriptors and Graph Representation}


\author[1]{\fnm{Dan} \sur{Luo}}

\author[1]{\fnm{Jinyu} \sur{Zhou}}

\author[1]{\fnm{Le} \sur{Xu}}

\author*[2]{\fnm{Sisi} \sur{Yuan}}\email{syuan4@charlotte.edu}

\author*[1,3]{\fnm{Xuan} \sur{Lin}}\email{jack\_lin@xtu.edu.cn}

\affil[1*]{\orgdiv{School of Computer Science}, 
           \orgname{Xiangtan University}, 
           \orgaddress{\city{Xiangtan}, 
                       \postcode{411105}, 
                       \country{China}}}
\affil[2*]{\orgdiv{Department of Bioinformatics and Genomics}, 
           \orgname{the University of North Carolina at Charlotte}, 
           \orgaddress{\city{Charlotte, NC}, 
                       \postcode{28223}, 
                       \country{USA}}}
\affil[3]{\orgdiv{Key Laboratory of Intelligent Computing and Information Processing of Ministry of Education},                 \orgname{Xiangtan University}, 
          \orgaddress{\city{Xiangtan}, 
                       \postcode{411105}, 
                       \country{China}}}
                       



\abstract{\textbf{Motivation:} Predicting drug–target binding affinity (DTA) is essential for identifying potential therapeutic candidates in drug discovery. However, most existing models rely heavily on static protein structures, often overlooking the dynamic nature of proteins, which is crucial for capturing conformational flexibility that will be benefical for protein binding interactions.

\textbf{Methods:} We introduce DynamicDTA, an innovative deep learning framework that incorporates static and dynamic protein features to enhance DTA prediction. The proposed DynamicDTA takes three types of inputs, including drug sequence, protein sequence, and dynamic descriptors. A molecular graph representation of the drug sequence is generated and subsequently processed through graph convolutional network, while the protein sequence is encoded using dilated convolutions. Dynamic descriptors, such as root mean square fluctuation, are processed through a multi-layer perceptron. These embedding features are fused with static protein features using cross-attention, and a tensor fusion network integrates all three modalities for DTA prediction.

\textbf{Results:} Extensive experiments on three datasets demonstrate that DynamicDTA achieves by at least 3.4\% improvement in $e_{\mathrm{RMSE}}$ score with comparison to seven state-of-the-art baseline methods. Additionally, predicting novel drugs for \textit{Human Immunodeficiency Virus Type 1} and visualizing the docking complexes further demonstrates the reliability and biological relevance of DynamicDTA.

\textbf{Availability and implementation:} The source code is publicly available and can be accessed at \url{https://github.com/shmily-ld/DynamicDTA}.

\section*{Graphical Abstract}
\begin{center}
\includegraphics[width=1\textwidth]{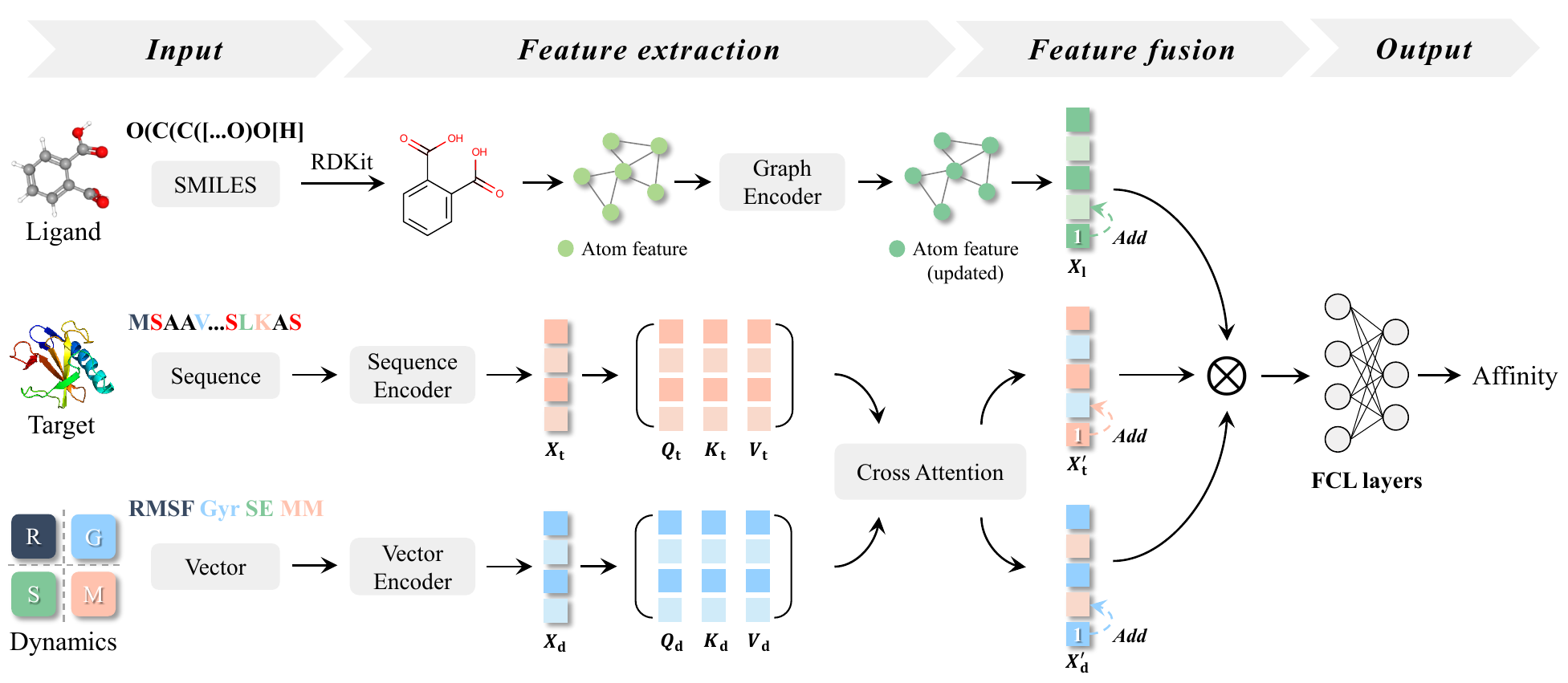}
\end{center}
}


\maketitle


\section{Introduction}\label{sec1}

An initial step in the drug discovery pipeline \cite{le2023predicting} is to identify molecules that bind to specific protein targets with high affinity and specificity, which can be further developed into drug-like molecules \cite{zhang2024prediction}. Drug-target binding affinity (DTA) is a crucial metric that quantifies the strength of interaction between a pair of drug and target protein, playing a pivotal role in the efficacy and specificity of potential therapeutic compounds \cite{tang2024mff}. Accurate prediction of DTA is essential in drug discovery \cite{d2023deep, sadybekov2023computational}, as it facilitates virtual screening, lead optimization, and toxicity mitigation \cite{tanoori2021drug}. By prioritizing compounds with high affinity, guiding structural modifications, and predicting off-target interactions, DTA prediction reduces biochemical validation costs and accelerates the identification of druggable candidates, ultimately improving the efficiency of the drug discovery \cite{pei2023breaking, zhu2024drug}.

Traditional experimental methods for measuring binding affinity, such as isothermal titration calorimetry and surface plasmon resonance, provide precise measurements but are often time-consuming, resource-intensive, and impractical for large-scale screening. To overcome these limitations, computational methods have emerged as efficient alternatives \cite{pan2022deep, zhang2023multi}. These methods leverage data from public repositories like Kiba \cite{tang2014making} and Davis \cite{davis2011comprehensive} to predict binding affinities by leveraging the molecular and structural characteristics of drugs and proteins. Among these computational strategies, deep learning has become a powerful approach \cite{vo2023improved}, owing to its unparalleled capacity to model complex, non-linear relationships between input features \cite{stepniewska2018development}.

Within the domain of deep learning-based DTA prediction, methods can be broadly categorized into sequence-based and structure-based approaches. Sequence-based models utilize the primary sequence information of drugs and proteins, providing a high-level abstraction of molecular characteristics. A seminal work in this domain, DeepDTA \cite{Ozturk2018DeepDTA},  demonstrated the utility of convolutional neural network (CNN) in capturing local sequence patterns and interactions, achieving substantial improvements over traditional methods. WideDTA \cite{ozturk2019widedta} is further proposed to integrate four sources of text-based information. AttentionDTA \cite{zhao2022attentiondta} combines an attention mechanism with the prediction of binding affinity, enhancing the extraction of relevant features from both drugs and protein sequences, resulting in a comprehensive deep learning-driven framework. Unlike static word embedding, DEAttentionDTA \cite{chen2024deattentiondta} integrates dynamic word encoding with multi-head self-attention, enabling multi-scale feature interaction between drugs and target proteins. In contrast, structure-based models capitalize on the three-dimensional conformational information of drugs and proteins to provide a more granular representation of molecular interactions \cite{li2024predicting}. Such as, GraphDTA \cite{nguyen2021graphdta} employs graph neural network (GNN) to model molecular graphs, thereby facilitating a more nuanced understanding of molecular interactions. Similarly, DGraphDTA \cite{jiang2020drug} represents proteins as contact maps and utilizes GNN to simultaneously learn the features of both drugs and proteins, enhancing the accuracy of interaction predictions. ImageDTA \cite{han2024imagedta} focuses on the 2D structure of drugs, treating drug 2D representations as “images” and processing them with multiscale 2D-CNNs for better interpretability and performance. The known protein sequences \cite{jumper2021highly} number in the billions, while the currently identified protein structures \cite{wwpdb2019protein} represent only a small portion. In practice, obtaining the accurate 3D structure of the drug-target complex or even the protein itself is often challenging \cite{chu2022hierarchical, wang2024chl}. Consequently, our work exclusively employs molecular graphs, bypassing the use of protein graphs.

A common limitation of most existing approaches is their reliance on static features of proteins, which fail to capture the inherently dynamic nature of these biomolecules. Proteins are not rigid entities, and they undergo continuous structural fluctuations, including changes in atomic positions and binding postures, driven by their biological environment and functional role. These dynamic characteristics are critical for understanding key processes such as molecular recognition, allosteric regulation, and binding interactions \cite{lu2024dynamicbind}. By affecting the accessibility, shape, and compatibility of binding sites, protein dynamics directly influence the specificity and strength of drug-target interactions \cite{wang2024prediction}. Extracting dynamic features allows models to capture this vital information, leading to more accurate predictions and a deeper understanding of protein function, ultimately facilitating more effective drug discovery.

To address this challenge, we propose DynamicDTA, an innovative framework designed to predict DTA by incorporating dynamic protein features derived from molecular dynamics (MD) simulations \cite{hashemzadeh2019study}. By integrating time-dependent structural descriptors, such as root mean square fluctuation (RMSF) \cite{liu2024dynamic}, DynamicDTA provides a more realistic representation of protein behavior. These dynamic features allow the model to capture temporal variations in protein conformations, thereby enhancing its ability to predict binding affinity. DynamicDTA introduces several innovations to advance the state-of-the-art methods in DTA prediction. First, it incorporates dynamic features derived from molecular dynamics simulations into the model input. Second, it uses a cross-attention mechanism to learn both static and dynamic protein features, enabling the model to capture their interactions more effectively. Additionally, a tensor fusion network (TFN) \cite{zadeh2017tensor} is employed to integrate multi-modal information from both drugs and proteins, ensuring comprehensive feature fusion. Comprehensive experiments have shown that DynamicDTA consistently outperforms seven state-of-the-art methods. Overall, our contributions are summarized as follows:
\begin{itemize}
\item We introduce DynamicDTA, an innovative framework that incorporates dynamic protein features derived from MD simulations, offering a more precise representation of protein behavior for predicting binding affinity.
\item We leverage a cross-attention mechanism to capture both static and dynamic protein features, while a TFN seamlessly integrates multi-modal data from both drugs and proteins, enabling the model to effectively combine diverse feature representations and enhance feature fusion.
\item Comparison experiments on three datasets have shown that DynamicDTA outperforms seven state-of-the-art methods. Furthermore, a case study predicting potential drugs for \textit{Human Immunodeficiency Virus Type 1} showcases the model's potential in accelerating drug discovery.
\end{itemize}

\section{Materials and Methods}\label{sec2}
\subsection{Datasets}\label{subsec2}
The protein dynamic descriptors are sourced from ATLAS dataset \cite{vander2024atlas}, which currently includes 1,938 proteins and provides four types of protein dynamic features from MD simulations:
\begin{enumerate}[label=(\arabic*)]
\item Avg.RMSF (average RMSF): This descriptor measures the fluctuation of each atom or residue from its average position during MD simulations. Specifically, it quantifies the root mean square deviation of atomic positions over time, providing insights into the flexibility of individual residues or regions within a protein. Typically, residues with high Avg.RMSF values are found in flexible regions such as loops or unstructured coils, which may be involved in essential protein functions like substrate binding, enzymatic regulation, or protein-protein interactions \cite{song2024accurate}. Conversely, residues with low Avg.RMSF values are often located in rigid regions such as $\alpha$-helices and $\beta$-sheets, which maintain structural stability and play a crucial role in preserving the overall protein conformation \cite{henzler2007dynamic}. In drug design, regions with high RMSF values could serve as potential drug targets due to their dynamic adaptability in drug-target interactions \cite{ghahremanian2022molecular}.
\item Avg.Gyr (average gyration radius): Gyr measures the overall compactness of the protein structure by calculating the average distance between the protein's atoms and its center of mass. Avg.Gyr over the simulation period provides insight into the stability and folding of the protein. A smaller Avg.Gyr suggests a more compact structure, which may indicate an active conformation or a stable binding state with other molecules. Conversely, a larger Avg.Gyr may suggest a more extended protein conformation, potentially associated with dynamic processes such as folding, unfolding, or dissociation from other molecules \cite{lobanov2008radius}. Changes in the radius of gyration can reflect dynamic interactions and conformational transitions within the structure, which are essential for understanding their functional mechanisms.
\item Div.SE (minimum TM-score between start and final conformations): This descriptor quantitatively evaluates the structural divergence between the initial and final conformations of a protein by employing TM-score \cite{zhang2005tm}, which is a widely used metric for assessing the similarity of protein structures. A high Div.SE value indicates minimal structural changes during the simulation, suggesting that the protein maintains a stable conformation, which may imply high structural rigidity under physiological conditions, allowing it to perform its biological function effectively. Conversely, a low Div.SE value suggests significant conformational changes, which could be associated with the protein's dynamic functionality, such as structural adaptations required for substrate binding or catalytic processes in enzymatic reactions \cite{lindorff2011fast, raval2012refinement}.
\item Div.MM (minimum TM-score between most divergent conformations): This descriptor assesses the structural diversity of a protein by measuring TM-score between the most divergent conformations observed during the simulation. It captures the protein's flexibility by identifying how much its structure changes between its most extreme conformations. A high Div.MM value suggests that, while the protein undergoes conformational changes, its overall structure remains similar, indicating flexibility without drastic rearrangements. Conversely, a low Div.MM value suggests significant conformational shifts, potentially linked to protein folding, allosteric regulation, or molecular interactions \cite{rivalta2012allosteric}.
\end{enumerate}

The affinity data are obtained from BindingDB \cite{liu2007bindingdb}, a comprehensive database of experimentally measured binding affinities, which contains 2.9 million entries across 1.3 million compounds and 9,400 targets. However, some data points for \( K_{\mathrm{i}} \), \( K_{\mathrm{d}} \), and \( \mathrm{IC}_{\mathrm{50}} \), which are commonly used metrics to measure binding affinity \cite{hua2023mfr}, were missing in the dataset. To address this issue, we split the data into three subsets based on different affinity measures. Additionally, to ensure that the affinity values fall within a suitable range for modeling, we applied a negative logarithmic transformation. Taking \( K_\mathrm{d}\) as an example, the transformation is applied using the following formula:
\begin{equation}
K_\mathrm{d}^{\prime} = -\log_{10}\left(\frac{K_\mathrm{d}}{1 \times 10^9}\right) \ 
\end{equation}
After the negative logarithmic transformation, some of the resulting values were negative. These values were removed, as they do not align with the expected range of affinity values. Finally, the values for \( K_\mathrm{d}^{\prime} \) range from 0.3 to 11.9, \( K_\mathrm{i}^{\prime} \) from 0.0 to 14.2, and \( \mathrm{IC}_{50}^{\prime} \) from 0.0 to 12.6.

To integrate the protein dynamic features with the affinity data, we perform a matching process based on PDB ID. As a result, three final datasets were crated, named as: \textit{K$_\mathrm{d}$}$^*$, \textit{K$_\mathrm{i}$}$^*$, and IC$_{50}$$^*$. Table \ref{tab:datasets} displays a summary of the key details of the three datasets.
\begin{table}[!t]
\centering
\caption{Summary of the datasets.}\label{tab:datasets} 
\renewcommand{\arraystretch}{1.5} 
\begin{tabular}{lccc}
\toprule
Dataset & Targets & Ligands & Binding Entities \\
\midrule
\textit{K}\(_\mathrm{d}\)$^{*}$  & 63  & 1,609  & 2,896  \\
\textit{K}\(_\mathrm{i}\)$^{*}$  & 74  & 18,872 & 27,106 \\
IC\(_\mathrm{50}\)$^{*}$ & 136 & 45,525 & 86,236 \\
Kiba$^{*}$ & 4 & 1,597 & 3,708 \\
\botrule
\end{tabular}
\end{table}

\begin{figure*}[!t]
\centering
\includegraphics[width=1\linewidth]{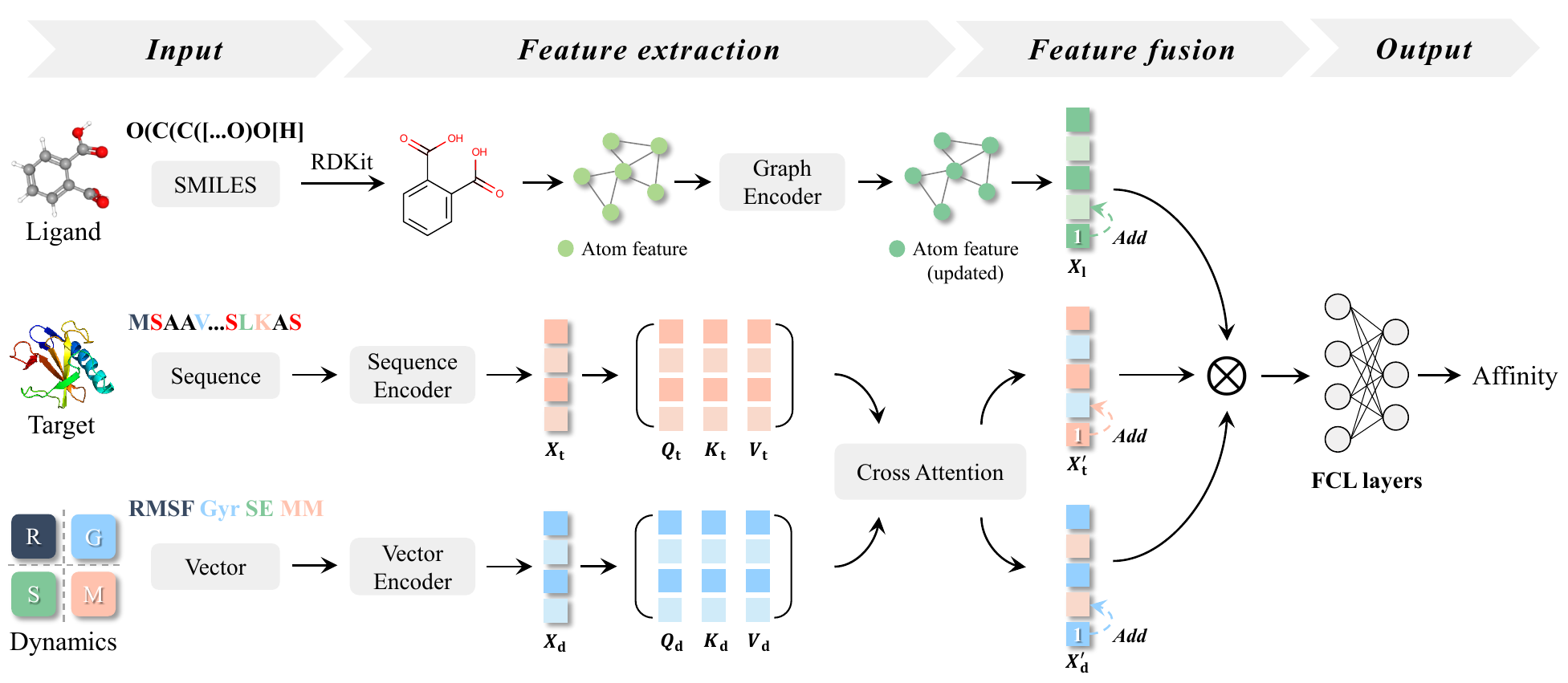}
\caption{The architecture of DynamicDTA. The framework integrates drug molecular graph, target sequences and dynamic descriptors, and a cross-attention mechanism to extract meaningful representations. TFN is employed to effectively fuse the extracted features for accurate drug–target binding affinity prediction.}
\label{fig:1}
\end{figure*}

We evaluated the model using five-fold cross-validation, as in AttentionDTA \cite{zhao2022attentiondta}. Each dataset was split into five parts, using one for testing and the rest for training in each fold. The final result were averaged across all folds. To further assess the model's generalization ability, we additionally processed Kiba dataset \cite{tang2014making} following the same procedure and ensuring that no proteins or drugs overlap with those in other datasets, resulting in a new dataset named Kiba$^*$. This external dataset was used to confirm the model’s ability to generalize to unseen data, providing further validation of its robustness.

\subsection{DynamicDTA Framework}\label{subsec2}
As depicted in Figure \ref{fig:1}, the DynamicDTA framework consists of four key components: input representation, feature extraction, feature fusion, and output block. In the following sections, we will describe each part of the framework in detail.

\subsubsection{Input Representation}\label{subsubsec2}
The DynamicDTA model takes three types of input data: ligands, targets, and dynamics. In what follows, we provide a detailed description of each input representation.

\noindent\textit{\textbf{Ligand representation.}} Ligands in this study are represented using the simplified molecular input line entry system (SMILES), a standardized and compact notation for encoding molecular structures as linear text strings \cite{weininger1988smiles}. To utilize the structural and chemical information captured by SMILES, we convert these strings into graph-based representations. The process begins by parsing each SMILES string into a molecular graph using the open-source chemical informatics software RDKit \cite{bento2020open}. In this graph, atoms serve as nodes, while chemical bonds act as edges. Each atom is represented  by a feature vector that encapsulates critical atomic properties, including atomic number, valence, and aromaticit. These feature vectors are normalized to maintain numerical stability across different molecular graphs. Bond information is extracted as pairs of indices indicating whether there is a bond between two atoms in the graph.

\noindent\textit{\textbf{Target representation.}} For target protein representation, each protein sequence is converted into a numerical vector using a predefined mapping of amino acid characters to integers. Each amino acid residue is assigned a unique integer, and characters not found in the mapping are assigned a default value of 0. To address the variation in protein sequence lengths, we define a fixed maximum length of 1000. Sequences longer than this threshold are truncated, while shorter sequences are padded with 0. This standardization ensures that all protein sequences have the same length, enabling efficient batch processing and maintaining consistent input dimensions for the model. 

\noindent\textit{\textbf{Dynamic representation.}} In this study, we focus on four key dynamic features to represent the dynamic nature of proteins: Avg.RMSF, Avg.Gyr, Div.SE, and Div.MM. To standardize these dynamic descriptors and make them comparable across proteins, we perform min-max normalization on each feature. Specifically, for a feature \( x \), the normalization is defined as
\begin{equation}
x_{\mathrm{norm}} = \frac{x - x_{\mathrm{min}}}{x_{\mathrm{max}} - x_{\mathrm{min}}} \
\end{equation}
where \( x_{\mathrm{min}} \) and \( x_{\mathrm{max}} \) represent the minimum and maximum values of feature \( x \) across the dataset. After normalization, the four dynamic features are concatenated into a single 4-dimensional vector. This vector serves as a comprehensive representation of the protein's dynamic behavior, capturing key aspects such as flexibility, structural transitions, and potential binding site changes, which are crucial for understanding drug-target interactions. The integration of these dynamic features improves the model's capacity to predict binding affinities by accounting for the protein’s conformational flexibility and structural transitions, making this approach a significant innovation in DTA prediction.

\subsubsection{Feature Extraction}\label{subsubsec2}
\noindent\textit{\textbf{Graph Encoder.}}  To extract meaningful features from the graph representation of a ligand, we utilize a graph convolutional network (GCN) to learn a graph-level representation that represents its molecular architecture. Given a ligand represented by graph \( G = (V, E) \), where \( V \) is the set of \( N \) nodes, each node has an associated feature vector \( \boldsymbol{x}_i \in \mathbb{R}^C \), and \( E \) is the set of edges described by the adjacency matrix \( \boldsymbol{A} \in \mathbb{R}^{N \times N} \). GCN uses the node feature matrix \( \boldsymbol{X} \) and the adjacency matrix \( \boldsymbol{A} \) as inputs. The propagation rule for GCN is given by
\begin{equation}
\boldsymbol{H}^{(l+1)} = \sigma \left( \tilde{\boldsymbol{D}}^{-1/2} \tilde{\boldsymbol{A}} \tilde{\boldsymbol{D}}^{-1/2} \boldsymbol{H}^{(l)} \boldsymbol{W}^{(l)} \right) \
\end{equation}
where \( \tilde{\boldsymbol{A}} = \boldsymbol{A} + \boldsymbol{I}_N \) is the adjacency matrix with self-connections, \( \tilde{\boldsymbol{D}} \) is the degree matrix of \( \tilde{\boldsymbol{A}} \), \( \boldsymbol{W}^{(l)} \) is the learnable weight matrix at layer \( l \), and \( \sigma \) is a non-linear activation function. After several layers, a global max pooling operation is applied to obtain a final feature vector of ligand \(\boldsymbol{X}_{\mathrm{l}}\).

\noindent\textit{\textbf{Sequence Encoder.}} Previous studies have shown that dilated convolution \cite{yu2015multi} is an effective technique for capturing multiscale contextual information by expanding the receptive field through different dilation rates \cite{jin2023capla}. Inspired by this, we apply dilated 1D convolutions to model long-range intramolecular interactions in protein sequences. While a standard 1D convolution operates on adjacent elements of the sequence with a fixed receptive field, the dilated 1D convolution increases the spacing between the filter elements, allowing it to effectively capture dependencies over long distances in the sequence without increasing the number of parameters. The dilated convolution operation for a given input sequence \( S = \{\boldsymbol{s}_1, \boldsymbol{s}_2, \dots, \boldsymbol{s}_N\} \), where \( \boldsymbol{s}_i \) represents the feature vector of amino acid \( i \), is performed as follows:
\begin{equation}
\boldsymbol{h}^{(l)}_i = \sigma \left( \sum_{j=1}^{K} \boldsymbol{w}_j^{(l)} \boldsymbol{s}_{i + r_j} + \boldsymbol{b}^{(l)} \right) \ 
\end{equation}
where \( \boldsymbol{w}_j^{(l)} \) are the weights of the convolutional filter at layer \( l \), \( r_j \) denotes the dilation rate for the \( j \)-th filter, \( K \) is the filter size, and \( \boldsymbol{b}^{(l)} \) is the bias term. Similar to the ligand representation, after passing through the dilated convolutions, the final feature vector of target \( \boldsymbol{X}_\mathrm{t} \) is obtained by applying a global max pooling operation.

\noindent\textit{\textbf{Vector Encoder.}} The dynamic features of each protein are represented as a normalized 4-dimensional vector \( \boldsymbol{V}_\mathrm{d}\). To extract high-level representations, the normalized vector \( \boldsymbol{V}_\mathrm{d}\) is passed through a multi-layer perceptron (MLP). The transformation at each MLP layer is defined as
\begin{equation}
\boldsymbol{X}_\mathrm{d}^{(l+1)} = \sigma \left( \boldsymbol{W}^{(l)} \boldsymbol{X}_\mathrm{d}^{(l)} + \boldsymbol{b}^{(l)} \right) \ 
\end{equation}
where \( \boldsymbol{X}_\mathrm{d}^{(0)} = \boldsymbol{V}_\mathrm{d} \) is the input vector, \( \boldsymbol{W}^{(l)} \) and \( \boldsymbol{b}^{(l)} \) represent the weight matrix and bias vector for the \( l \)-th layer, respectively. The final output \( \boldsymbol{X}_\mathrm{d} \) is a compact feature vector that captures the essential information from the dynamic descriptors.

\noindent\textit{\textbf{Cross Attention Mechanism.}} To effectively integrate complementary information between the target sequence vector \( \boldsymbol{X}_\mathrm{t} \) and the dynamic vector \( \boldsymbol{X}_\mathrm{d} \), we employ a multi-head cross attention mechanism, which captures bidirectional interactions, enabling \( \boldsymbol{X}_\mathrm{t} \) to utilize information from \( \boldsymbol{X}_\mathrm{d} \), and \( \boldsymbol{X}_\mathrm{d} \) to incorporate information from \( \boldsymbol{X}_\mathrm{t} \). Initially, the target and dynamic vectors are linearly transformed to compute their respective query, key, and value matrices for each head:
\begin{equation}
\boldsymbol{Q}_\mathrm{t}^{(i)} = \boldsymbol{W_Q}^{\mathrm{t},(i)} \boldsymbol{X}_\mathrm{t}
\end{equation}
\begin{equation}
\boldsymbol{K}_\mathrm{d}^{(i)} = \boldsymbol{W_K}^{\mathrm{d},(i)} \boldsymbol{X}_\mathrm{d}
\end{equation}
\begin{equation}
\boldsymbol{V}_\mathrm{d}^{(i)} = \boldsymbol{W_V}^{\mathrm{d},(i)} \boldsymbol{X}_\mathrm{d}
\end{equation}
where each head \( i = 1, \dots, \mathrm{H} \), where \( \mathrm{H} \) is the number of attention heads, and \( \boldsymbol{W_Q}^{\mathrm{t},(i)} \), \( \boldsymbol{W_K}^{\mathrm{d},(i)} \), and \( \boldsymbol{W_V}^{\mathrm{d},(i)} \) are learnable weight matrices. The attention computation for the \( i \)-th head is given by
\begin{equation}
\boldsymbol{A}_\mathrm{t}^{(i)} = \mathrm{Softmax}\left(\frac{\boldsymbol{Q}_\mathrm{t}^{(i)} (\boldsymbol{K}_\mathrm{d}^{(i)})^\top}{\sqrt{d_{\boldsymbol{K}}}}\right) {\boldsymbol{V}}_\mathrm{d}^{(i)} \ 
\end{equation}
where \( d_{\boldsymbol{K}} \) is the dimension of the key matrices. The multi-head outputs are concatenated along the feature dimension and then passed through a learned linear transformation:
\begin{equation}
\boldsymbol{X}_\mathrm{t}^{\prime} = \mathrm{Concat}(\boldsymbol{A}_\mathrm{t}^{(1)}, \dots, \boldsymbol{A}_\mathrm{t}^{(\mathrm{H})}) \cdot \boldsymbol{W}_\mathrm{t}^{\mathrm{O}} \ 
\end{equation}
where \( \boldsymbol{W}_\mathrm{t}^{\mathrm{O}} \) is the output projection matrix. Similarly, to enable the dynamic vector \( \boldsymbol{X}_\mathrm{d} \) to attend to the target vector \( \boldsymbol{X}_\mathrm{t} \), the corresponding query, key, and value matrices for each head are computed as
\begin{equation}
\boldsymbol{Q}_\mathrm{d}^{(i)} = \boldsymbol{W_Q}^{\mathrm{d},(i)} \boldsymbol{X}_\mathrm{d}
\end{equation}
\begin{equation}
\boldsymbol{K}_\mathrm{t}^{(i)} = \boldsymbol{W_K}^{\mathrm{t},(i)} \boldsymbol{X}_\mathrm{t}
\end{equation}
\begin{equation}
\boldsymbol{V}_\mathrm{t}^{(i)} = \boldsymbol{W_V}^{\mathrm{t},(i)} \boldsymbol{X}_\mathrm{t}
\end{equation}
The attention computation for the \( i \)-th head is
\begin{equation}
\boldsymbol{A}_\mathrm{d}^{(i)} = \mathrm{Softmax}\left(\frac{\boldsymbol{Q}_\mathrm{d}^{(i)} (\boldsymbol{K}_\mathrm{t}^{(i)})^\top}{\sqrt{d_{\boldsymbol{K}}}}\right) {\boldsymbol{V}}_\mathrm{t}^{(i)} \ 
\end{equation}
The multi-head outputs are concatenated along the feature dimension and then passed through a learned linear transformation:
\begin{equation}
\boldsymbol{X}_\mathrm{d}^{\prime} = \mathrm{Concat}(\boldsymbol{A}_\mathrm{d}^{(1)}, \dots, \boldsymbol{A}_\mathrm{d}^{(\mathrm{H})}) \cdot \boldsymbol{W}_\mathrm{d}^{\mathrm{O}} \ 
\end{equation}
The model effectively fuses complementary information from both vectors using the multi-head cross-attention mechanism, attending to each other in parallel and enriching their representations with contextual dependencies. This results in more informative and context-aware feature representations for both the target sequence and dynamic features.

\subsubsection{Feature Fusion}\label{subsec2}
We use TFN \cite{zadeh2017tensor} to integrate the features \( \boldsymbol{X}_\mathrm{t}^{\prime} \), \( \boldsymbol{X}_\mathrm{d}^{\prime} \), and \( \boldsymbol{X}_\mathrm{l} \) from the target, dynamics, and ligand, respectively. Specifically, we augment each modality's representation by adding an extra constant dimension with a value of 1. The extended representations are given by
\begin{equation}
\tilde{\boldsymbol{X}_\mathrm{l}} = 
\begin{bmatrix}
\boldsymbol{X}_\mathrm{l} \\
1
\end{bmatrix} \ ,
\quad
\tilde{\boldsymbol{X}_\mathrm{t}} = 
\begin{bmatrix}
\boldsymbol{X}_\mathrm{t}^{\prime} \\
1
\end{bmatrix} \ ,
\quad
\tilde{\boldsymbol{X}_\mathrm{d}} = 
\begin{bmatrix}
\boldsymbol{X}_\mathrm{d}^{\prime} \\
1
\end{bmatrix} \ 
\quad
\end{equation}
Next, we fuse the information from the three extended feature vectors, \( \tilde{\boldsymbol{X}_\mathrm{l}} \), \( \tilde{\boldsymbol{X}_\mathrm{t}} \), and \( \tilde{\boldsymbol{X}_\mathrm{d}} \), by computing their outer product:
\begin{equation}
\boldsymbol{X}_\mathrm{f} = \boldsymbol{W}_\mathrm{f} \cdot \big( \tilde{\boldsymbol{X}_\mathrm{t}} \otimes \tilde{\boldsymbol{X}_\mathrm{d}} \otimes \tilde{\boldsymbol{X}_\mathrm{l}} \big) + \boldsymbol{b}_\mathrm{f} \
\end{equation}
where $\otimes$ denotes the outer product operation, $ \tilde{\boldsymbol{X}_\mathrm{t}} \otimes \tilde{\boldsymbol{X}_\mathrm{d}} \otimes \tilde{\boldsymbol{X}_\mathrm{l}}$ is the higher-order tensor that captures interactions across unimodal, bimodal, and trimodal features, and $\boldsymbol{W}_\mathrm{f}$ and $\boldsymbol{b}_\mathrm{f}$ are learnable parameters. $\boldsymbol{X}_\mathrm{f}$ represents the final fused feature. TFN explicitly captures the complex interactions between ligand, target and dynamic features, facilitating the seamless integration of multimodal information into a comprehensive, high-dimensional feature representation.

\subsubsection{Prediction and Training Module}\label{subsec2}
The prediction module employs fully connected layers (FCLs) to process the fused feature vector $\boldsymbol{X}_\mathrm{f}$ obtained from TFN. ReLU is used as the activation function in each layer to capture complex nonlinear patterns, leading to the prediction of the final binding affinity score $\hat{y}$. DTA prediction is framed as a regression problem \cite{zhang2022predicting}. The model is optimized to minimize the mean squared error (MSE) loss function, which is defined as
\begin{equation}
   \mathcal{L} = \frac{1}{\mathrm{M}} \sum_{i=1}^{\mathrm{M}} \left( y_i - \hat{y}_i \right)^2 \ 
\end{equation}
where \( y_i \) denotes the actual binding affinity of the \( i \)-th sample, \( \hat{y}_i \) represents the predicted binding affinity, while \( \mathrm{M} \) represents the total number of samples.

\section{Result}\label{sec3}
In this section, we introduce and evaluate the performance of model with comparison to baseline methods, followed by a case study that demonstrates its practical application.

\subsection{Experimental Settings}\label{subsec2}
The DynamicDTA model was implemented using PyTorch and trained on a Linux server with 12 CPUs and an Nvidia GeForce RTX 4090 GPU with 24 GB of VRAM. We set the number of epochs to 1000, the batch size to 512, and the learning rate to $5\times10^{-4}$, respectively. These hyperparameters were chosen to ensure efficient training and convergence of the model. Additionally, we set the number of GCN hidden layers to 3, the attention heads to 4, the embedding dimension to 64 and the dropout rate to 0.2 to prevent overfitting, respectively. Other detailed model parameters can be found in the source code. The Adam optimizer \cite{kingma2014adam} was used to update the model parameters, and MSE was employed as the loss function. To ensure robust evaluation, five-fold cross-validation was performed across all datasets.

\subsection{Baseline Methods}\label{subsec2}
To assess the performance of our model, we compared the proposed DynamicDTA with seven baseline methods for DTA prediction, including two traditional machine learning models implemented with scikit-learn library and five state-of-the-art deep learning-based approaches. The deep learning baselines were implemented using the source code provided in the original papers, while the traditional machine learning models were built using standard implementations, without any additional tuning. The seven models used for comparison are the following:
\begin{itemize}
\item \textbf{Linear Regression} employs label encoding on drug SMILES and protein sequences, using concatenated vectors for linear affinity prediction.
\item \textbf{Decision Tree} applies the same label encoding followed by decision tree regression with feature space partitioning.
\item \textbf{DeepDTA} \cite{Ozturk2018DeepDTA} employs CNN to process protein sequences and drug SMILES representations for DTA.
\item \textbf{GraphDTA} \cite{nguyen2021graphdta} employs GNN to extract and learn molecular graph structural representations, combined with protein sequence embeddings for binding affinity prediction.
\item \textbf{AttentionDTA} \cite{zhao2022attentiondta} employs an attention-based architecture to enhance the extraction of relevant features from drug SMILES and protein sequences.
\item \textbf{DEAttentionDTA} \cite{chen2024deattentiondta} uses a 1D CNN for dynamic word embedding and captures interactions between drugs and proteins features by integrating a self-attention mechanism.
\item \textbf{ImageDTA} \cite{han2024imagedta} leverages a multiscale 2D CNN to process SMILES-encoded molecules and enhancs interpretability through convolutional kernel size selection.
\end{itemize}

\subsection{Evaluation Metrics}\label{subsec2}
To quantitatively evaluate the predictive performance of our DTA model, we employ two widely used metrics, the root mean square error (RMSE, denoted by $e_{\mathrm{RMSE}}$) as well as the Pearson correlation coefficient (\textit{R}). The specific formulas are as follows:
\begin{equation}
e_{\mathrm{RMSE}} = \sqrt{\frac{1}{\mathrm{M}} \sum_{i=1}^{\mathrm{M}} (y_i - \hat{y}_i)^2} \ 
\end{equation}
\begin{equation}
R = \frac{\sum\limits_{i=1}^{\mathrm{M}} (y_i - \bar{y})(\hat{y}_i - \bar{\hat{y}})}{\sqrt{\sum\limits_{i=1}^{\mathrm{M}} (y_i - \bar{y})^2 \sum\limits_{i=1}^{\mathrm{M}} (\hat{y}_i - \bar{\hat{y}})^2}} \ 
\end{equation}
where $e_{\mathrm{RMSE}}$ measures the average magnitude of prediction errors, with lower values indicating better prediction accuracy. \textit{R} measures the linear relationship between the predicted and true values, with higher values indicating a stronger correlations between them.

\subsection{Comparison Results}\label{subsec2}
\begin{table*}[h]
\caption{Comparison performance of DynamicDTA and baseline methods on three datasets.}\label{tab:Comparison}
\renewcommand{\arraystretch}{1.5} 
\footnotesize
\begin{tabular*}{\textwidth}{@{\extracolsep\fill}lcccccc}
\toprule
\multirow{2.5}{*}{Model} & \multicolumn{2}{@{}c@{}}{${\mathrm{IC}_{50}}^{*}$} & \multicolumn{2}{@{}c@{}}{${K_{\mathrm{d}}}^{*}$} & \multicolumn{2}{@{}c@{}}{${K_{\mathrm{i}}}^{*}$} \\
\cmidrule{2-3} \cmidrule{4-5} \cmidrule{6-7}
 & $e_{\mathrm{RMSE}}$(std) \(\downarrow\) & \textit{R}(std) \(\uparrow\) & $e_{\mathrm{RMSE}}$(std) \(\downarrow\) & \textit{R}(std) \(\uparrow\) & $e_{\mathrm{RMSE}}$(std) \(\downarrow\) & \textit{R}(std) \(\uparrow\) \\
\midrule
Linear Regression & 1.293(0.035) & 0.412(0.005) & 1.648(0.131) & 0.322(0.067) & 1.374(0.020) & 0.538(0.010) \\
Decision Tree & 1.190(0.049) & 0.662(0.007) & 1.369(0.078) & 0.716(0.032) & 1.321(0.018) & 0.586(0.008) \\
DeepDTA & 0.792(0.089) & 0.887(0.002) & 1.312(0.139) & 0.754(0.065) & 0.904(0.019) & 0.839(0.004) \\
GraphDTA & 0.625(0.008) & \underline{0.918(0.002)} & \underline{1.029(0.054)} & \underline{0.816(0.015)} & \underline{0.792(0.017)} & \underline{0.866(0.011)} \\
AttentionDTA & 0.702(0.020) & 0.897(0.007) & 1.119(0.042) & 0.786(0.014) & 0.861(0.021) & 0.850(0.005) \\
DEAttentionDTA & 0.846(0.053) & 0.847(0.020) & 1.157(0.189) & 0.759(0.089) & 1.058(0.112) & 0.757(0.065) \\
ImageDTA & \textbf{0.600(0.053)} & 0.853(0.010) & 1.595(0.157) & 0.780(0.010) & 0.896(0.030) & 0.804(0.003) \\
\textbf{DynamicDTA} & \underline{0.611(0.006)} & \textbf{0.923(0.002)} & \textbf{1.007(0.050)} & \textbf{0.830(0.012)} & \textbf{0.727(0.017)} & \textbf{0.892(0.003)} \\
\botrule
\end{tabular*}
\begin{threeparttable}
\begin{tablenotes}
\item\textit{Note: the best results are represented in bold, and the second-best results are underlined.}
\end{tablenotes}
\end{threeparttable}
\end{table*}

\begin{figure*}[!ht]
\centering
\includegraphics[width=1\linewidth]{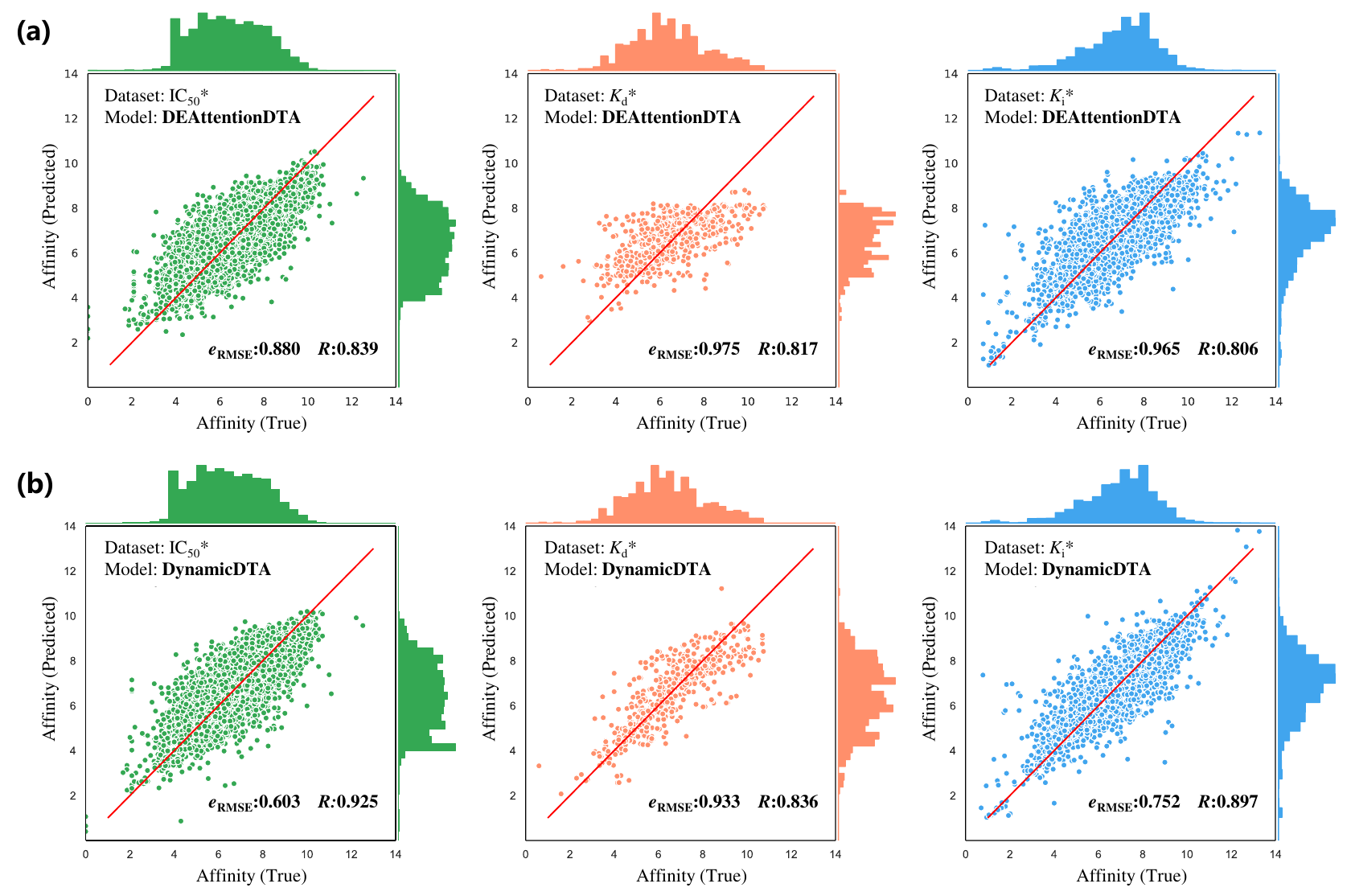}
\caption{The performance of DEAttentionDTA (a) and DynamicDTA (b) on three datasets for the prediction of DTA.}
\label{fig:2}
\end{figure*}

Table \ref{tab:Comparison} summarizes the comparison performance of DynamicDTA with seven baseline methods on three datasets. The results demonstrate that our model consistently outperforms all baseline methods. On the ${\textit{K}_\mathrm{d}}^{*}$ dataset, DynamicDTA achieves a notable improvement, surpassing GraphDTA, the second-best model, by 2.2\% in $e_{\mathrm{RMSE}}$ and 1.4\% in \textit{R}. Similarly, on the ${\textit{K}_\mathrm{i}}^{*}$ dataset, DynamicDTA again outperformes the second-best method, with improvement of 6.5\% in $e_{\mathrm{RMSE}}$ and 2.6\% in \textit{R}. However, on the ${\mathrm{IC}_\mathrm{50}}^{*}$ dataset, while DynamicDTA achieves the highest \textit{R}, it falls slightly behind ImageDTA in $e_{\mathrm{RMSE}}$. This discrepancy may stem from the nature of the \(\mathrm{IC}_{50}\) metric, which focuses more on a drug's inhibitory effects rather than directly correlating with binding affinity \cite{barbet2019equilibrium, lagunin2018comparison}. Furthermore, ImageDTA places greater emphasis on drug-specific features, which could explain its slight advantage in this specific metric. Traditional machine learning models perform poorly compared to deep learning-based methods like DeepDTA due to their reliance on handcrafted features, which may not fully capture the complex interactions between drugs and proteins. In contrast, deep learning models automatically extract hierarchical features, leading to better generalization and performance for DTA prediction.

\begin{table*}[!ht]
\caption{Ablation study on three datasets.}\label{tab:Ablation}
\renewcommand{\arraystretch}{1.5} 
\footnotesize
\begin{tabular*}{\textwidth}{@{\extracolsep\fill}lcccccc}
\toprule%
\multirow{2.5}{*}{Model} & \multicolumn{2}{@{}c@{}}{${\mathrm{IC}_{50}}^{*}$} & \multicolumn{2}{@{}c@{}}{${K_{\mathrm{d}}}^{*}$} & \multicolumn{2}{@{}c@{}}{${K_{\mathrm{i}}}^{*}$} \\
\cmidrule{2-3} \cmidrule{4-5} \cmidrule{6-7}
 & $e_{\mathrm{RMSE}}$(std) \(\downarrow\) & \textit{R}(std) \(\uparrow\) & $e_{\mathrm{RMSE}}$(std) \(\downarrow\) & \textit{R}(std) \(\uparrow\) & $e_{\mathrm{RMSE}}$(std) \(\downarrow\) & \textit{R}(std) \(\uparrow\) \\
\midrule
w/o Dilated & 0.611(0.006) & 0.923(0.002) & 1.021(0.053) & 0.824(0.012) & 0.732(0.011) & 0.890(0.002) \\
w/o RMSF+Gyr  & 0.611(0.009) & 0.923(0.003) & 1.018(0.047) & 0.826(0.011) & 0.735(0.013) & 0.889(0.002) \\
w/o SE+MM & 0.614(0.007) & 0.920(0.001) & 1.041(0.048) & 0.820(0.012) & 0.760(0.015) & 0.890(0.004) \\
w/o RMSF+Gyr+SE & 0.613(0.007) & 0.922(0.002) & 1.023(0.045) & 0.825(0.011) & 0.742(0.016) & 0.887(0.005) \\
w/o Gyr+SE+MM & 0.613(0.005) & 0.922(0.001) & 1.031(0.051) & 0.821(0.015) & 0.751(0.019) & 0.884(0.005) \\
\textbf{DynamicDTA} & \textbf{0.611(0.006)} & \textbf{0.923(0.002)} & \textbf{1.007(0.050)} & \textbf{0.830(0.012)} & \textbf{0.727(0.017)} & \textbf{0.892(0.003)} \\
\botrule
\end{tabular*}
\begin{threeparttable}
\begin{tablenotes}
\item\textit{Note: the best results are represented in bold.}
\end{tablenotes}
\end{threeparttable}
\end{table*}

Figure \ref{fig:2} shows the predictive performance of DynamicDTA in comparison to DEAttentionDTA, using scatter plots of true versus predicted binding affinities. This comparison reveals that DynamicDTA demonstrates a tighter clustering of points along the diagonal, indicating more accurate predictions across a wider range of binding affinities. The difference in predictive performance between DynamicDTA and DEAttentionDTA stems from the unique features of each model. While DEAttentionDTA uses dynamic word embeddings and self-attention for linear protein and ligand sequences, DynamicDTA improves this by integrating graph-based ligand representations and dynamic protein features. This allows DynamicDTA to better capture the structural complexity and flexibility of drug-target interactions, with dynamic protein descriptors from MD simulations further enhancing its ability to model protein dynamics.

\subsection{Ablation Study}\label{subsec2}
In this section, we performed an ablation study to evaluate the contribution of each component in the model. Specifically, we analyzed the effect of dilated convolution, dynamic descriptors, and different feature fusion strategies. These analyses allow us to quantify the importance of each component and provide deeper insights into their influence on DTA prediction.

\subsubsection{Ablation on Dilated Convolution} 
We replaced dilated convolutions with standard convolutions to assess their role in feature extraction. In this experiment, w/o Dilated denotes the use of standard convolutions instead of dilated convolutions. The results of this ablation study are summarized in Table \ref{tab:Ablation}. The substitution of dilated convolutions with standard convolutions resulted in a decline in both $e_{\mathrm{RMSE}}$ and \textit{R} metrics, highlighting the importance of dilated convolutions in capturing multi-scale features for better predictive performance.

\subsubsection{Ablation on Dynamic Descriptors}
\begin{table*}[!h]
\caption{Comparison of feature fusion methods on three datasets.}\label{tab:Feature Fusion}
\renewcommand{\arraystretch}{1.5} 
\footnotesize
\begin{tabular*}{\textwidth}{@{\extracolsep\fill}lcccccc}
\toprule%
\multirow{2.5}{*}{Fusion Method} & \multicolumn{2}{@{}c@{}}{${\mathrm{IC}_{50}}^{*}$} & \multicolumn{2}{@{}c@{}}{${K_{\mathrm{d}}}^{*}$} & \multicolumn{2}{@{}c@{}}{${K_{\mathrm{i}}}^{*}$} \\
\cmidrule{2-3} \cmidrule{4-5} \cmidrule{6-7}
 & $e_{\mathrm{RMSE}}$(std) \(\downarrow\) & \textit{R}(std) \(\uparrow\) & $e_{\mathrm{RMSE}}$(std) \(\downarrow\) & \textit{R}(std) \(\uparrow\) & $e_{\mathrm{RMSE}}$(std) \(\downarrow\) & \textit{R}(std) \(\uparrow\) \\
\midrule
Concat & 0.618(0.007) & \underline{0.921(0.002)} & \underline{1.008(0.044)} & \underline{0.825(0.010)} & 0.763(0.023) & 0.880(0.005) \\
Sum  & \underline{0.616(0.010)} & 0.920(0.003) & 1.144(0.040) & 0.813(0.015) & 0.751(0.010) & \underline{0.883(0.012)}\\
Average & 0.618(0.005) & 0.919(0.002)  & 1.122(0.010) & 0.823(0.030) & 0.793(0.019) & 0.866(0.010) \\
Hadamard product & 0.620(0.006) & 0.920(0.005) & 1.115(0.023) & 0.821(0.020) & \underline{0.730(0.015)} & 0.877(0.007)\\
\textbf{Ours} & \textbf{0.611(0.006)} & \textbf{0.923(0.002)} & \textbf{1.007(0.050)} & \textbf{0.830(0.012)} & \textbf{0.727(0.017)} & \textbf{0.892(0.003)} \\
\botrule
\end{tabular*}
\begin{threeparttable}
\begin{tablenotes}
\item\textit{Note: the best results are represented in bold, and the second-best results are underlined.}
\end{tablenotes}
\end{threeparttable}
\end{table*}

We conducted a series of ablation experiments on dynamic protein descriptors to evaluate their impact on model performance. In these experiments, w/o RMSF+Gyr indicates the removal of RMSF and Gyr from DynamicDTA, while w/o SE+MM represents the exclusion of SE and MM. Similarly, w/o RMSF+Gyr+SE and w/o Gyr+SE+MM correspond to the removal of these respective sets of dynamic descriptors, allowing us to analyze their contributions to the model. Table \ref{tab:Ablation} summarizes the results of this ablation study. Removing dynamic protein features led to a decline in model performance, with the impact being more pronounced when multiple features were excluded. The removal of RMSF and Gyr alone resulted in only minor performance fluctuations, suggesting that RMSF, which captures residue-level flexibility, and Gyr, which reflects overall structural compactness, may be partially redundant. In contrast, when Gyr, SE, and MM were simultaneously removed, the model exhibited a significant decline in performance. This shows that SE, which represents the minimum TM-score between the initial and final conformations, and MM, which captures the minimum TM-score between the most structurally divergent conformations, together provide critical insights into large-scale conformational changes. The substantial performance drop upon their removal highlights the importance of these descriptors in capturing protein structural transitions and flexibility, which are crucial for accurate binding affinity prediction.

\begin{figure*}[!ht]
\centering
\includegraphics[width=1\linewidth]{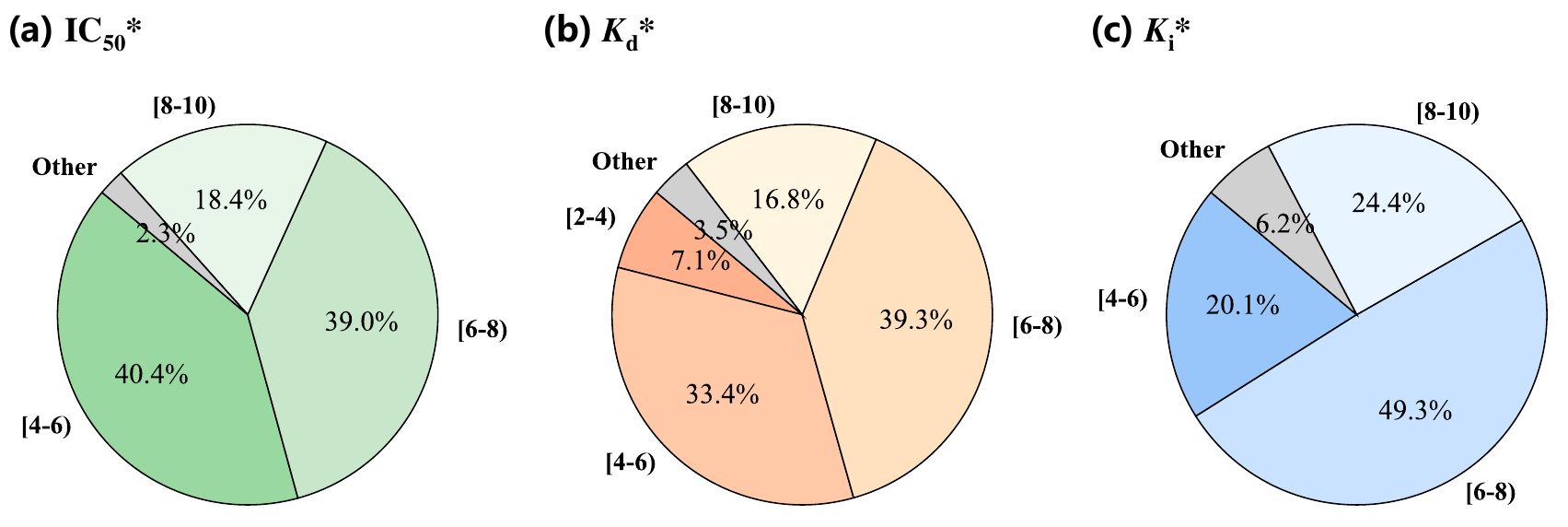}
\caption{Affinity distribution comparison across datasets: (a) \({\mathrm{IC}_{50}}^*\) dataset. (b) \({K_{\mathrm{d}}}^*\) dataset. (c) \({K_{\mathrm{i}}}^*\) dataset.}
\label{fig:3}
\end{figure*}

\subsubsection{Ablation on Feature Fusion}
To further evaluate the effectiveness of our feature fusion approach, we conducted additional experiments comparing different fusion methods, including Concat, Sum, Average, Hadamard product, and ours (TFN). In the Concat method, drug, target, and dynamic features are concatenated along the feature dimension. The Sum method aggregates these features by element-wise addition, while the Average method further normalizes the sum by dividing the number of modalities. The Hadamard product method performs element-wise multiplication of the feature representations. Each fusion method was applied before passing the combined features to the final prediction layer. As shown in Table \ref{tab:Feature Fusion}, the TFN-based fusion method achieves the best performance across all datasets, with the lowest $e_{\mathrm{RMSE}}$ and highest \textit{R}. Traditional fusion methods such as Concat and Sum perform slightly worse, suggesting that TFN better captures complex feature interactions for improved binding affinity prediction.  This improvement can be attributed to TFN’s ability to model higher-order correlations between features, which are often crucial for understanding intricate biomolecular interactions. Moreover, the enhanced performance across multiple datasets further demonstrates the robustness and generalization capability of the TFN approach.

\begin{figure*}
\centering
\includegraphics[width=1\linewidth]{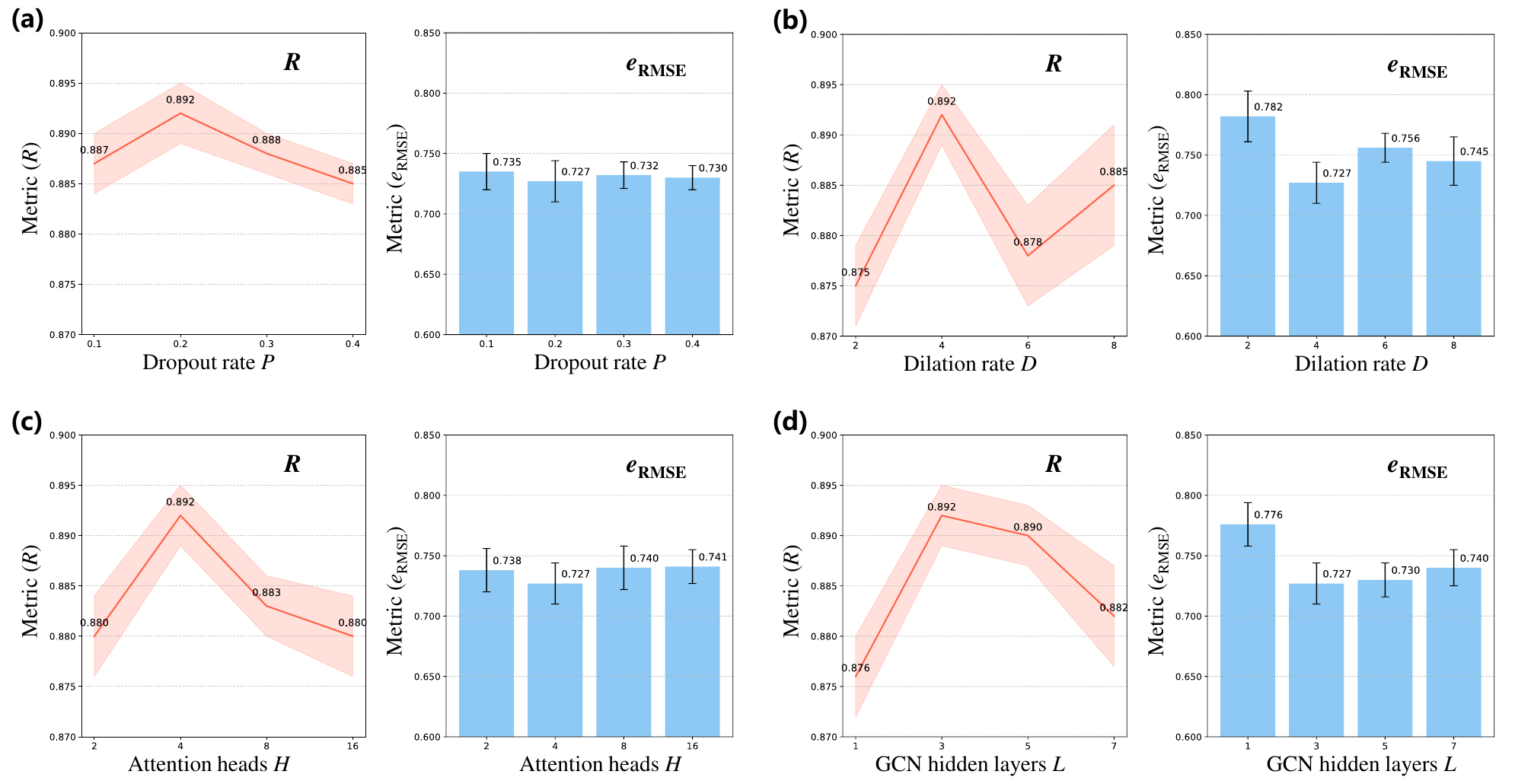}
\caption{Parameter sensitivity analysis in the \({K_{\mathrm{i}}}^*\) dataset: (a) Dropout rate \( P \). (b) Dilation rate \( D \). (c) Attention heads \( H \). (d) GCN hidden layers \( L \).}
\label{fig:4}
\end{figure*}

Interestingly, the performance improvement observed in the \({K_{\mathrm{i}}}^*\) dataset was more pronounced than in other datasets, both in the ablation and comparison experiments. This can be attributed to the inherent characteristics of the dataset. As shown in Figure \ref{fig:3}, the affinity distribution across the three datasets further highlights this difference. Specifically, the binding affinities in the \({K_{\mathrm{i}}}^*\) dataset are more concentrated, which simplifies the prediction task. This concentration enables the model to generalize better, making it more responsive to the contributions of each component.

\subsection{Parameter Sensitivity Analysis}\label{subsec2}

In this section, we systematically investigate the influence of four critical hyperparameters in our DynamicDTA model: the dropout rate \( P \) of the regularization, the dilation rate \( D \) of the dilated convolution, the number of attention heads \( H \) in the cross-attention layer, and the number of GCN hidden layers \( L \) for ligands.

We first examine the effect of the dropout rate \( P \), which controls the regularization strength to prevent overfitting. We evaluated \( P \) values from the set \( \{0.1, 0.2, 0.3, 0.4\} \). As shown in Figure \ref{fig:4}a, the model achieves the best performance when \( P = 0.2 \). A lower dropout rate may lead to overfitting, while a higher dropout rate may excessively reduce the model’s capacity.

Next, we analyze the dilation rate \( D \) in the dilated convolution module, considering values from \( \{2, 4, 6, 8\} \). Figure \ref{fig:4}b shows that the optimal performance is achieved at \( D = 4 \). A small dilation rate may limit the receptive field, making it difficult to capture long-range dependencies, whereas an excessively large dilation rate may dilute local information.

For the cross-attention layer, we evaluate different numbers of attention heads \( H \) from the set \( \{2, 4, 8, 16\} \). As illustrated in Figure \ref{fig:4}c, the model achieves optimal performance when \( H = 4 \). Increasing \( H \) can improve the model’s ability to capture diverse interactions, but an excessive number may result in overfitting or oversmoothing.

Finally, we investigate the effect of the number of GCN hidden layers \( L \) for ligand representation, testing values from \( \{1, 3, 5, 7\} \). Figure \ref{fig:4}d indicates that the best performance is achieved when \( L = 3 \). A shallow network may lack the capacity to extract complex patterns, while an overly deep network may suffer from gradient vanishing and overfitting.

\subsection{External Dataset for Validation}

To further evaluate the generalization capability of our DynamicDTA model, we apply the trained model directly to the Kiba$^*$ dataset without fine-tuning. Since we performed five-fold cross-validation, we tested the best-performing model from each fold on the Kiba$^*$ dataset, and the final performance is reported as the average of these results. Table \ref{tab:external} presents the performance comparison between our model and baseline methods in terms of $e_\mathrm{RMSE}$ and \textit{R}. The results show that DynamicDTA outperforms the baseline methods on the Kiba$^*$ dataset, achieving approximately a 4.3\% improvement in $e_{\mathrm{RMSE}}$ and a 13\% improvement in \textit{R} over the second-best model. Note that since no fine-tuning was performed on the Kiba$^*$ dataset, the absolute performance is not very high (e.g., $e_{\mathrm{RMSE}}$ is around 1.0 and \textit{R} is typically around 0.9 in comparison experiments, as shown in Table~\ref{tab:Comparison}). These results further highlight the better generalization ability of the DynamicDTA model, demonstrating its robustness and effectiveness across different datasets without requiring additional fine-tuning.

\begin{table}[!t]
\centering
\caption{Generalization performance comparison on the Kiba$^*$ Dataset.}\label{tab:external} 
\renewcommand{\arraystretch}{1.5} 
\begin{tabular}{lcc}
\toprule
Model & $e_{\mathrm{RMSE}}$(std) \(\downarrow\) & \textit{R}(std) \(\uparrow\)  \\
\midrule
Liner Regression & 6.615(0.020) & -0.056(0.008) \\
Decision Tree & 5.636(0.015) & -0.071(0.006) \\
DeepDTA  & 6.212(0.010)  & 0.072(0.003)  \\
GraphDTA  & 5.954(0.023) & \underline{0.176(0.007)} \\
AttentionDTA & \underline{5.323(0.015)} & 0.093(0.005) \\
DEAttentionDTA & 5.816(0.012) & -0.015(0.005) \\
ImageDTA & 5.599(0.008) & 0.010(0.004) \\
\textbf{DynamicDTA} & \textbf{5.090(0.013)} & \textbf{0.202(0.002)} \\
\botrule
\end{tabular}
\begin{tablenotes}
\item\textit{Note: the best results are represented in bold, and the second-best results are underlined.}
\end{tablenotes}
\end{table}

\subsection{Interpretation and Application}
In this section, we analyze the model's interpretability by visualizing the attention weights assigned to protein residues and conducting a case study to explore how the model makes predictions from a dynamic perspective.
\begin{figure*}[!ht]
\centering
\includegraphics[width=1\linewidth]{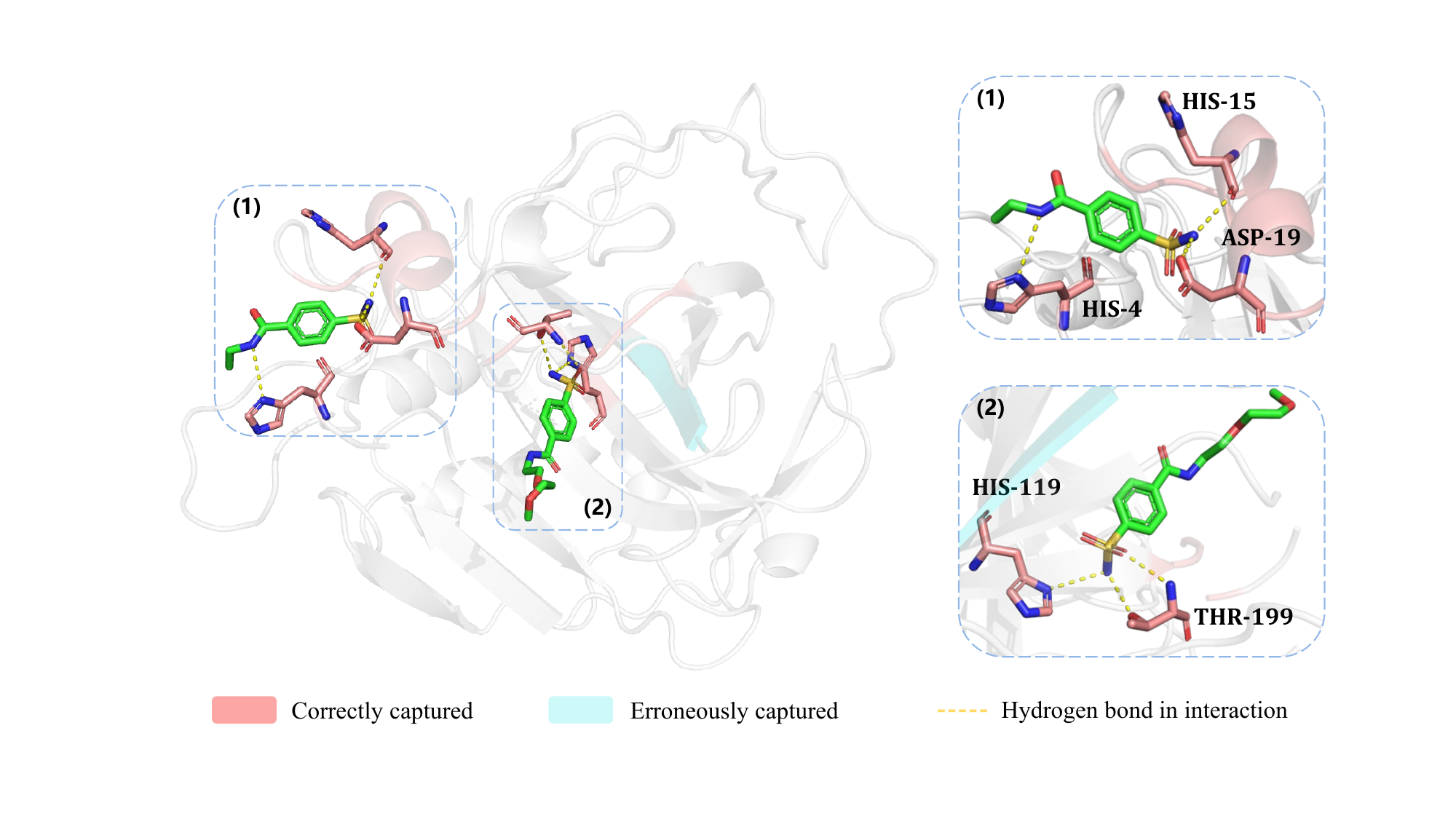}
\caption{Visualization of drug-target interactions in the 2FOS complex, which consists of two distinct ligand-binding regions. The model's attention weights for the top 20 residues are highlighted, where correctly identified binding residues are shown in red, and misidentified residues are shown in cyan. The two smaller panels on the right provide zoomed-in views of the two distinct ligand-binding regions from the left panel. These views are rotated to optimal angles to offer a clearer perspective of the binding interactions.}
\label{Interpretability analysis}
\end{figure*}

\subsubsection{Interpretability Analysis}
Inspired by NHGNN-DTA \cite{he2023nhgnn} that employs attention weight visualization for interpretability analysis, we conducted a similar analysis on the DynamicDTA model. Specifically, we examined the cross-attention weights assigned to protein residues, as our model applies the attention mechanism exclusively to protein features, integrating protein sequence features and dynamic descriptors. Since the drug features are processed without an attention mechanism, we visualized only the attention weights associated with protein residues to identify critical binding regions prioritized by the model during DTA prediction. For this analysis, the 2FOS drug-target complex from the RCSB Protein Data Bank (RCSB PDB) was selected as a representative case. This crystallographically resolved structure contains two distinct ligand-binding sites, providing an ideal framework for validating the model's interpretability.

As illustrated in Figure \ref{Interpretability analysis}, the 2FOS complex exhibits two ligand-binding pockets, labeled (1) and (2) in the schematic representation. The model accurately localized residues within these functional regions. In the attention map, the top 20 residues with the highest attention weights are color-coded: red denotes correctly identified binding residues, while cyan indicates residues with erroneously elevated attention weights. Notably, the high-attention regions (red) exhibit strong spatial overlap with experimentally validated binding sites, demonstrating the model's capability to effectively identify pharmacologically relevant interactions. A detailed examination of binding site (1) reveales that the model successfully recognizes three key interacting residues: HIS-4, HIS-15, and ASP-19 (highlighted in red). These residues are known from crystallographic studies to form hydrogen bonds and electrostatic interactions with the ligand, a finding consistent with the model's attention patterns. This alignment between computational predictions and experimental structural data reinforces the biological plausibility of DynamicDTA's decision-making process.

This analysis provides insights into how the DynamicDTA model focuses on key residues involved in ligand binding and highlights the model's ability to predict these critical interactions. The inclusion of attention weight visualization further strengthens the model's interpretability and aids in understanding the underlying mechanisms of drug-target interactions.

\subsubsection{Case Study}\label{subsec2}

\begin{table*}[!t]
\setlength{\tabcolsep}{27pt} 
\caption{The top 10 predicted potential drugs targeting HIV-1.}\label{tab:case_study} 
\renewcommand{\arraystretch}{1.5} 
\footnotesize
\begin{tabular}{lccc}
\toprule
\textbf{Rank} & \textbf{Drug Name} & \textbf{DrugBank ID} & \textbf{Evidence} \\ 
\midrule
1 & L-methionine (R)-S-oxide & DB02235 & Unconfirmed \\
2 & Procainamide & DB01035 & Unconfirmed \\
3 & \textbf{Oseltamivir} & \textbf{DB00198} & \textbf{PMID: 36067538} \\
4 & \textbf{Hypochlorite} & \textbf{DB11123} & \textbf{PMID: 10773730} \\
5 & \textbf{Tromethamine} & \textbf{DB03754} & \textbf{PMID: 34602806} \\
6 & \textbf{Isoleucine} & \textbf{DB00167} & \textbf{PMID: 34454514} \\
7 & Bendazac & DB13501 & Unconfirmed \\
8 & 3-Bromo-7-Nitroindazole & DB01997 & Unconfirmed \\
9 & \textbf{Deferiprone} & \textbf{DB08826} & \textbf{PMID: 27191165} \\
10 & \textbf{Formaldehyde} & \textbf{DB03843} & \textbf{PMID: 37632035} \\
\botrule
\end{tabular}
\begin{threeparttable}
\begin{tablenotes}
\item\textit{Note: the drugs with related evidence found in PubMed are represented in bold.}
\end{tablenotes}
\end{threeparttable}
\end{table*}

In case study, we applied the DynamicDTA model to predict potential drugs that target \textit{Human Immunodeficiency Virus Type 1} (HIV-1, PDB ID: 4JMU). We first extracted dynamic features of this target from our dataset and paired them with drug-target pairs from the DrugBank database. These pairs were subsequently input into the model to predict the affinity between each drug-target protein pair. After ranking the drugs based on their predicted affinity, we cross-checked the top candidates with PubMed\footnote{\url{https://pubmed.ncbi.nlm.nih.gov/}} to gather supporting evidence regarding their potential therapeutic effects against HIV-1. The results, presented in Table \ref{tab:case_study}, list the top 10 predicted drugs. These findings highlight the potential of our model in accelerating drug discovery.

\begin{figure*}[!t]
\centering
\includegraphics[width=1\linewidth]{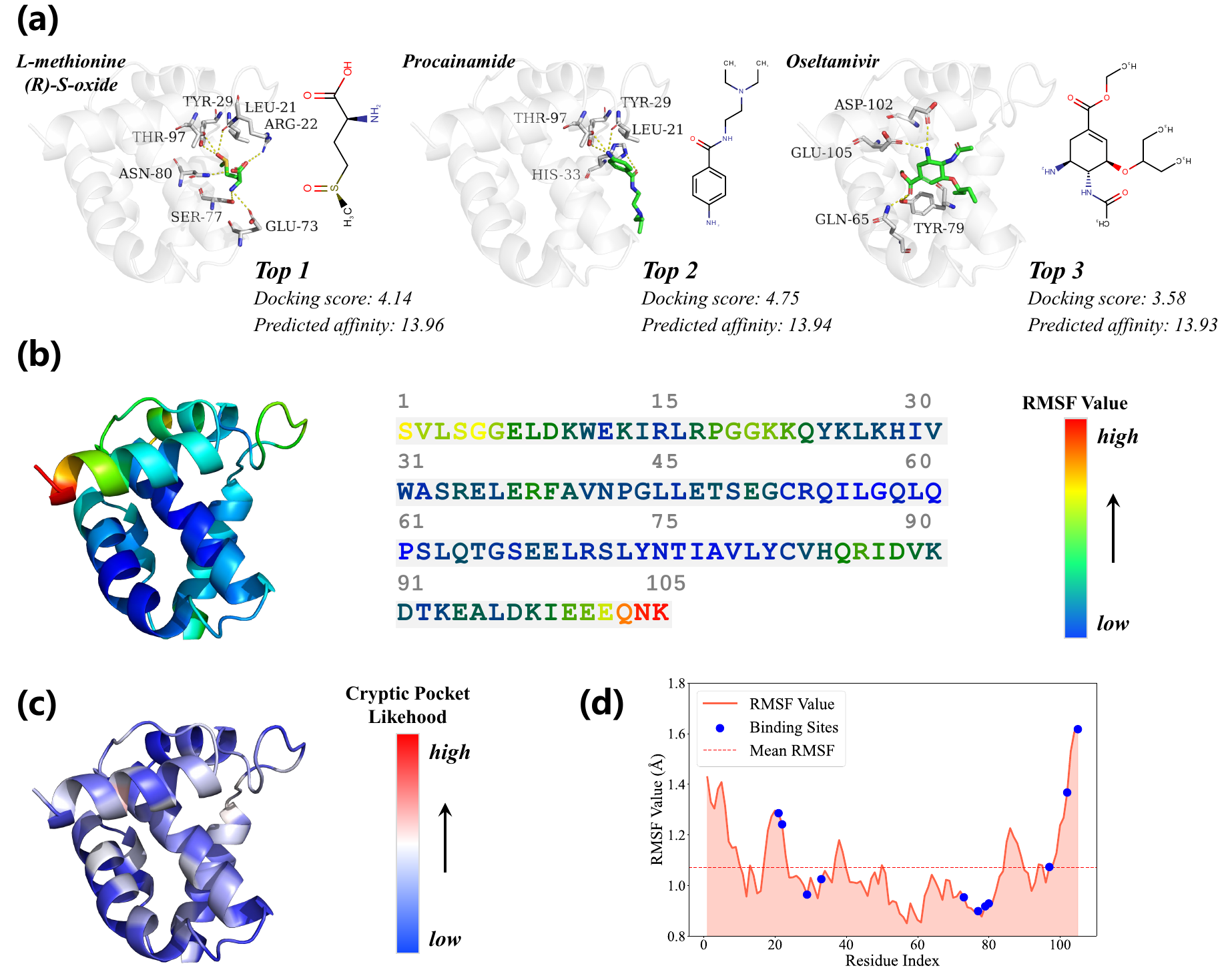}
\caption{Comprehensive analysis of drug-protein interactions and dynamic properties of the target protein. (a) Docking of the top three predicted drug candidates with the HIV-1 target protein, highlighting key binding residues. (b) Visualization of residue-level RMSF values mapped onto the 3D protein structure (left) and sequence (right), illustrating rigid and flexible regions. (c) PocketMiner-predicted likelihood of each residue contributing to a cryptic binding pocket, with higher values indicating potential transient binding sites. (d) Line plot of RMSF values along the residue index, with blue dots marking residues involved in drug binding, demonstrating the correlation between binding sites and dynamic properties.}
\label{fig:5}
\end{figure*}

To gain deeper insights into the interactions between the predicted drugs and the target protein, we performed a series of visualizations and analyses. First, we utilized the DeepMice\footnote{\url{http://www.deepmice.com/}} online molecular docking \cite{li2019overview} web server to dock the top three predicted candidate drugs with the target protein, generating corresponding drug-protein complexes. These complexes were subsequently visualized using PyMOL\footnote{\url{https://www.pymol.org/}}, as shown in Figure \ref{fig:5}a. These visualizations illustrate the residue regions involved in the interactions between each drug and the HIV-1 target. Notably, Figure \ref{fig:5}a presents both the docking score and the predicted binding affinity. It is important to emphasize that these two metrics reflect different aspects of the drug-target interaction. The docking score, shown in the figure, is already relatively high, indicating strong binding potential, while the predicted binding affinity provides a quantitative estimate of the actual interaction strength.

We further analyzed the dynamic properties of the target protein at the residue level. Specifically, we visualized the RMSF values of each residue, mapping the values to a color gradient to highlight regions of varying flexibility. As shown in Figure \ref{fig:5}b, the left panel presents the 3D structure of the protein visualized in PyMOL, while the right panel depicts the sequence-based visualization of RMSF values. Notably, some residues involved in drug binding, such as SER-77, exhibit low RMSF values, suggesting that these regions are rigid concave surfaces, which are commonly observed binding sites for small-molecule ligands. In contrast, residues like LEU-21 display higher RMSF values, indicating increased flexibility. We hypothesize that these high-RMSF regions may correspond to cryptic binding sites \cite{vskrhak2024cryptobench}, which are transient pockets that emerge due to protein conformational changes. To validate this hypothesis, we employed the PocketMiner \cite{meller2023predicting} model, a specialized tool for predicting the likelihood of individual residues contributing to cryptic pocket formation. The results, shown in Figure \ref{fig:5}c, illustrate a strong correlation between high RMSF values and an increased likelihood of forming cryptic pockets. However, it is worth noting that some residues with high RMSF values do not significantly contribute to pocket formation, likely due to their inherent instability, which prevents stable drug binding.

Furthermore, Figure \ref{fig:5}d presents a line plot of RMSF values across the residue index, with blue dots highlighting the binding residues identified in the molecular docking study (Figure \ref{fig:5}a). This visualization clearly demonstrates that certain binding residues reside in low-RMSF regions, reinforcing the idea that stable, rigid concave surfaces often serve as primary binding sites. Meanwhile, other binding residues are located in high-RMSF regions, corresponding to cryptic pockets.

Overall, this case study demonstrates that our DynamicDTA model effectively captures the dynamic properties of target proteins, offering valuable insights into potential drug-binding regions and accelerating the drug discovery process.

\section{Discussion and Conclusion}\label{sec4}
Accurately predicting drug-target binding affinity (DTA) is crucial in drug discovery. Existing methods often rely on static protein and drug representations, neglecting the dynamic nature of molecular interactions. To address this, we propose DynamicDTA, a framework that integrates dynamic protein features with graph-based drug representations to model complex drug-target interactions. Specifically, a cross-attention mechanism is introduced to capture both static and dynamic protein features, while a TFN seamlessly integrates multi-modal data from both drugs and proteins. Comparison experiments on three datasets demonstrate that DynamicDTA outperforms state-of-the-art methods. Moreover,  a case study predicting potential drugs for HIV-1 further illustrates the model's capability in expediting the drug discovery process. However, we acknowledge several limitations in this study. First, the reliance on MD-derived dynamic descriptors may limit the model’s applicability to real world where MD simulations are computationally expensive, and comprehensive MD data are not always available. Second, inconsistencies in experimental binding affinity measurements across datasets could introduce noise, affecting model performance and generalizability. Addressing these issues will be crucial for future improvements. To overcome these limitations, in future work, we plan to explore generative AI-based approaches, such as diffusion models \cite{corso2022diffdock, schneuing2024structure}, to simulate MD data for proteins that lack experimental structures, reducing the computational burden while preserving essential dynamic information. 

\backmatter

\bmhead{Acknowledgements}
Thank the editor and anonymous reviewers for their valuable comments and suggestions. The work is supported in part by the National Natural Science Foundation of China (Grant No. 62202413), in part by the Research Foundation of Education Bureau of Hunan Province, China (Grant No. 23B0129), and in part by the Student Science and Technology Innovation Activity Plan of Hunan Province (Grant No. S202410530287).

\bmhead{Data availability}
Data openly available in a public repository. The protein dynamic descriptors data are available at \url{https://doi.org/10.1093/nar/gkad1084}, and the affinity data can be found in \url{https://www.bindingdb.org/} and \url{https://www.kaggle.com/datasets/christang0002/davis-and-kiba}. All codes are available at \url{https://github.com/shmily-ld/DynamicDTA} or from the corresponding author upon reasonable request.

\section*{Declarations}
The authors have no competing interests to declare that are relevant to the content of this article.

\
\bibliography{sn-bibliography}

\begin{thebibliography}{56}
\providecommand{\natexlab}[1]{#1}
\providecommand{\url}[1]{{#1}}
\providecommand{\urlprefix}{URL }
\providecommand{\doi}[1]{\url{https://doi.org/#1}}
\providecommand{\eprint}[2][]{\url{#2}}
 \bibcommenthead

\bibitem[{Le(2023)}]{le2023predicting}
Le NQK (2023) Predicting emerging drug interactions using gnns. Nature Computational Science 3(12):1007--1008. \doi{10.1038/s43588-023-00555-7}

\bibitem[{Zhang et~al.(2024)Zhang, Liu, Cheng, Wang, and Chen}]{zhang2024prediction}
Zhang H, Liu X, Cheng W et~al (2024) Prediction of drug-target binding affinity based on deep learning models. Computers in Biology and Medicine p 108435. \doi{10.1016/j.compbiomed.2024.108435}

\bibitem[{Tang et~al.(2024)Tang, Ma, Yang, and Li}]{tang2024mff}
Tang X, Ma W, Yang M et~al (2024) Mff-dta: Multi-scale feature fusion for drug-target affinity prediction. Methods 231:1--7. \doi{10.1016/j.ymeth.2024.08.008}

\bibitem[{D’Souza et~al.(2023)D’Souza, Prema, Balaji, and Shah}]{d2023deep}
D’Souza S, Prema K, Balaji S et~al (2023) Deep learning-based modeling of drug--target interaction prediction incorporating binding site information of proteins. Interdisciplinary Sciences: Computational Life Sciences 15(2):306--315. \doi{10.1007/s12539-023-00557-z}

\bibitem[{Sadybekov and Katritch(2023)}]{sadybekov2023computational}
Sadybekov AV, Katritch V (2023) Computational approaches streamlining drug discovery. Nature 616(7958):673--685. \doi{10.1038/s41586-023-05905-z}

\bibitem[{Tanoori et~al.(2021)Tanoori, Jahromi, and Mansoori}]{tanoori2021drug}
Tanoori B, Jahromi MZ, Mansoori EG (2021) Drug-target continuous binding affinity prediction using multiple sources of information. Expert Systems with Applications 186:115810. \doi{10.1016/j.eswa.2021.115810}

\bibitem[{Pei et~al.(2023)Pei, Wu, Zhu, Xia, Xie, Qin, Liu, Liu, and Yan}]{pei2023breaking}
Pei Q, Wu L, Zhu J et~al (2023) Breaking the barriers of data scarcity in drug--target affinity prediction. Briefings in Bioinformatics 24(6):bbad386. \doi{10.1093/bib/bbad386}

\bibitem[{Zhu et~al.(2024)Zhu, Zheng, Qi, Gong, Li, Mazur, Cong, and Gao}]{zhu2024drug}
Zhu Z, Zheng X, Qi G et~al (2024) Drug--target binding affinity prediction model based on multi-scale diffusion and interactive learning. Expert Systems with Applications 255:124647. \doi{10.1016/j.eswa.2024.124647}

\bibitem[{Pan et~al.(2022)Pan, Lin, Cao, Zeng, Yu, He, Nussinov, and Cheng}]{pan2022deep}
Pan X, Lin X, Cao D et~al (2022) Deep learning for drug repurposing: Methods, databases, and applications. Wiley interdisciplinary reviews: Computational molecular science 12(4):e1597. \doi{10.1002/wcms.1597}

\bibitem[{Zhang et~al.(2023)Zhang, Li, Wang, Xu, and Gu}]{zhang2023multi}
Zhang X, Li Y, Wang J et~al (2023) A multi-perspective model for protein--ligand-binding affinity prediction. Interdisciplinary Sciences: Computational Life Sciences 15(4):696--709. \doi{10.1007/s12539-023-00582-y}

\bibitem[{Tang et~al.(2014)Tang, Szwajda, Shakyawar, Xu, Hintsanen, Wennerberg, and Aittokallio}]{tang2014making}
Tang J, Szwajda A, Shakyawar S et~al (2014) Making sense of large-scale kinase inhibitor bioactivity data sets: a comparative and integrative analysis. Journal of Chemical Information and Modeling 54(3):735--743. \doi{10.1021/ci400709d}

\bibitem[{Davis et~al.(2011)Davis, Hunt, Herrgard, Ciceri, Wodicka, Pallares, Hocker, Treiber, and Zarrinkar}]{davis2011comprehensive}
Davis MI, Hunt JP, Herrgard S et~al (2011) Comprehensive analysis of kinase inhibitor selectivity. Nature biotechnology 29(11):1046--1051. \doi{10.1038/nbt.1990}

\bibitem[{Vo et~al.(2023)Vo, Nguyen, and Le}]{vo2023improved}
Vo TH, Nguyen NTK, Le NQK (2023) Improved prediction of drug-drug interactions using ensemble deep neural networks. Medicine in Drug Discovery 17:100149. \doi{10.1016/j.medidd.2022.100149}

\bibitem[{Stepniewska-Dziubinska et~al.(2018)Stepniewska-Dziubinska, Zielenkiewicz, and Siedlecki}]{stepniewska2018development}
Stepniewska-Dziubinska MM, Zielenkiewicz P, Siedlecki P (2018) Development and evaluation of a deep learning model for protein--ligand binding affinity prediction. Bioinformatics 34(21):3666--3674. \doi{10.1093/bioinformatics/bty374}

\bibitem[{{\"O}zt{\"u}rk et~al.(2018){\"O}zt{\"u}rk, {\"O}zg{\"u}r, and Ozkirimli}]{Ozturk2018DeepDTA}
{\"O}zt{\"u}rk H, {\"O}zg{\"u}r A, Ozkirimli E (2018) Deepdta: deep drug--target binding affinity prediction. Bioinformatics 34(17):i821--i829. \doi{10.1093/bioinformatics/bty593}

\bibitem[{{\"O}zt{\"u}rk et~al.(2019){\"O}zt{\"u}rk, Ozkirimli, and {\"O}zg{\"u}r}]{ozturk2019widedta}
{\"O}zt{\"u}rk H, Ozkirimli E, {\"O}zg{\"u}r A (2019) Widedta: prediction of drug-target binding affinity. arXiv preprint arXiv:190204166 \doi{10.48550/arXiv.1902.04166}

\bibitem[{Zhao et~al.(2022)Zhao, Duan, Yang, Cheng, Li, and Wang}]{zhao2022attentiondta}
Zhao Q, Duan G, Yang M et~al (2022) Attentiondta: Drug--target binding affinity prediction by sequence-based deep learning with attention mechanism. IEEE/ACM transactions on computational biology and bioinformatics 20(2):852--863. \doi{10.1109/TCBB.2022.3170365}

\bibitem[{Chen et~al.(2024)Chen, Huang, Shen, Zhang, Xu, Yang, Xie, Yan, and Yan}]{chen2024deattentiondta}
Chen X, Huang J, Shen T et~al (2024) Deattentiondta: Protein-ligand binding affinity prediction based on dynamic embedding and self-attention. Bioinformatics p btae319. \doi{10.1093/bioinformatics/btae319}

\bibitem[{Li et~al.(2024)Li, Yuan, and Zhang}]{li2024predicting}
Li G, Yuan Y, Zhang R (2024) Predicting protein--ligand binding affinity using fusion model of spatial-temporal graph neural network and 3d structure-based complex graph. Interdisciplinary Sciences: Computational Life Sciences pp 1--20. \doi{10.1007/s12539-024-00644-9}

\bibitem[{Nguyen et~al.(2021)Nguyen, Le, Quinn, Nguyen, Le, and Venkatesh}]{nguyen2021graphdta}
Nguyen T, Le H, Quinn TP et~al (2021) Graphdta: predicting drug--target binding affinity with graph neural networks. Bioinformatics 37(8):1140--1147. \doi{10.1093/bioinformatics/btaa921}

\bibitem[{Jiang et~al.(2020)Jiang, Li, Zhang, Wang, Wang, Yuan, and Wei}]{jiang2020drug}
Jiang M, Li Z, Zhang S et~al (2020) Drug--target affinity prediction using graph neural network and contact maps. RSC advances 10(35):20701--20712. \doi{10.1039/D0RA02297G}

\bibitem[{Han et~al.(2024)Han, Kang, and Guo}]{han2024imagedta}
Han L, Kang L, Guo Q (2024) Imagedta: A simple model for drug--target binding affinity prediction. ACS omega 9(26):28485--28493. \doi{10.1021/acsomega.4c02308}

\bibitem[{Jumper et~al.(2021)Jumper, Evans, Pritzel, Green, Figurnov, Ronneberger, Tunyasuvunakool, Bates, {\v{Z}}{\'\i}dek, Potapenko et~al.}]{jumper2021highly}
Jumper J, Evans R, Pritzel A et~al (2021) Highly accurate protein structure prediction with alphafold. nature 596(7873):583--589. \doi{10.1038/s41586-021-03819-2}

\bibitem[{wwp(2019)}]{wwpdb2019protein}
 (2019) Protein data bank: the single global archive for 3d macromolecular structure data. Nucleic acids research 47(D1):D520--D528. \doi{10.1093/nar/gky949}

\bibitem[{Chu et~al.(2022)Chu, Huang, Fu, Quan, Zhou, Liu, and Zhang}]{chu2022hierarchical}
Chu Z, Huang F, Fu H et~al (2022) Hierarchical graph representation learning for the prediction of drug-target binding affinity. Information Sciences 613:507--523. \doi{10.1016/j.ins.2022.09.043}

\bibitem[{Wang et~al.(2024)Wang, Liu, Zhang, Zhang, Song, Zhang, and Pang}]{wang2024chl}
Wang S, Liu Y, Zhang Y et~al (2024) Chl-dti: A novel high--low order information convergence framework for effective drug--target interaction prediction. Interdisciplinary Sciences: Computational Life Sciences pp 1--11. \doi{10.1007/s12539-024-00608-z}

\bibitem[{Lu et~al.(2024)Lu, Zhang, Huang, Zhang, Jia, Wang, Shi, Li, Wolynes, and Zheng}]{lu2024dynamicbind}
Lu W, Zhang J, Huang W et~al (2024) Dynamicbind: Predicting ligand-specific protein-ligand complex structure with a deep equivariant generative model. Nature Communications 15(1):1071. \doi{10.1038/s41467-024-45461-2}

\bibitem[{Wang(2024)}]{wang2024prediction}
Wang H (2024) Prediction of protein--ligand binding affinity via deep learning models. Briefings in Bioinformatics 25(2):bbae081. \doi{10.1093/bib/bbae081}

\bibitem[{Hashemzadeh et~al.(2019)Hashemzadeh, Ramezani, and Rafii-Tabar}]{hashemzadeh2019study}
Hashemzadeh S, Ramezani F, Rafii-Tabar H (2019) Study of molecular mechanism of the interaction between mek1/2 and trametinib with docking and molecular dynamic simulation. Interdisciplinary Sciences: Computational Life Sciences 11:115--124. \doi{10.1007/s12539-018-0305-4}

\bibitem[{Liu et~al.(2024)Liu, Wang, Cai, Wang, Kuang, Cheng, Zhang, Su, Tang, Cao et~al.}]{liu2024dynamic}
Liu C, Wang J, Cai Z et~al (2024) Dynamic pdb: A new dataset and a se (3) model extension by integrating dynamic behaviors and physical properties in protein structures. arXiv preprint arXiv:240812413 \doi{10.48550/arXiv.2408.12413}

\bibitem[{Zadeh et~al.(2017)Zadeh, Chen, Poria, Cambria, and Morency}]{zadeh2017tensor}
Zadeh A, Chen M, Poria S et~al (2017) Tensor fusion network for multimodal sentiment analysis. arXiv preprint arXiv:170707250 \doi{10.48550/arXiv.1707.07250}

\bibitem[{Vander~Meersche et~al.(2024)Vander~Meersche, Cretin, Gheeraert, Gelly, and Galochkina}]{vander2024atlas}
Vander~Meersche Y, Cretin G, Gheeraert A et~al (2024) Atlas: protein flexibility description from atomistic molecular dynamics simulations. Nucleic acids research 52(D1):D384--D392. \doi{10.1093/nar/gkad1084}

\bibitem[{Song et~al.(2024)Song, Bao, Feng, Huang, Zhang, Gao, and Han}]{song2024accurate}
Song X, Bao L, Feng C et~al (2024) Accurate prediction of protein structural flexibility by deep learning integrating intricate atomic structures and cryo-em density information. Nature Communications 15(1):5538. \doi{10.1038/s41467-024-49858-x}

\bibitem[{Henzler-Wildman and Kern(2007)}]{henzler2007dynamic}
Henzler-Wildman K, Kern D (2007) Dynamic personalities of proteins. Nature 450(7172):964--972. \doi{10.1038/nature06522}

\bibitem[{Ghahremanian et~al.(2022)Ghahremanian, Rashidi, Raeisi, and Toghraie}]{ghahremanian2022molecular}
Ghahremanian S, Rashidi MM, Raeisi K et~al (2022) Molecular dynamics simulation approach for discovering potential inhibitors against sars-cov-2: A structural review. Journal of molecular liquids 354:118901. \doi{10.1016/j.molliq.2022.118901}

\bibitem[{Lobanov et~al.(2008)Lobanov, Bogatyreva, and Galzitskaya}]{lobanov2008radius}
Lobanov MY, Bogatyreva N, Galzitskaya O (2008) Radius of gyration as an indicator of protein structure compactness. Molecular Biology 42:623--628. \doi{10.1134/S0026893308040195}

\bibitem[{Zhang and Skolnick(2005)}]{zhang2005tm}
Zhang Y, Skolnick J (2005) Tm-align: a protein structure alignment algorithm based on the tm-score. Nucleic acids research 33(7):2302--2309. \doi{10.1093/nar/gki524}

\bibitem[{Lindorff-Larsen et~al.(2011)Lindorff-Larsen, Piana, Dror, and Shaw}]{lindorff2011fast}
Lindorff-Larsen K, Piana S, Dror RO et~al (2011) How fast-folding proteins fold. Science 334(6055):517--520. \doi{10.1126/science.1208351}

\bibitem[{Raval et~al.(2012)Raval, Piana, Eastwood, Dror, and Shaw}]{raval2012refinement}
Raval A, Piana S, Eastwood MP et~al (2012) Refinement of protein structure homology models via long, all-atom molecular dynamics simulations. Proteins: Structure, Function, and Bioinformatics 80(8):2071--2079. \doi{10.1002/prot.24098}

\bibitem[{Rivalta et~al.(2012)Rivalta, Sultan, Lee, Manley, Loria, and Batista}]{rivalta2012allosteric}
Rivalta I, Sultan MM, Lee NS et~al (2012) Allosteric pathways in imidazole glycerol phosphate synthase. Proceedings of the National Academy of Sciences 109(22):E1428--E1436. \doi{10.1073/pnas.1120536109}

\bibitem[{Liu et~al.(2007)Liu, Lin, Wen, Jorissen, and Gilson}]{liu2007bindingdb}
Liu T, Lin Y, Wen X et~al (2007) Bindingdb: a web-accessible database of experimentally determined protein--ligand binding affinities. Nucleic acids research 35(suppl\_1):D198--D201. \doi{10.1093/nar/gkl999}

\bibitem[{Hua et~al.(2023)Hua, Song, Feng, and Wu}]{hua2023mfr}
Hua Y, Song X, Feng Z et~al (2023) Mfr-dta: a multi-functional and robust model for predicting drug--target binding affinity and region. Bioinformatics 39(2):btad056. \doi{10.1093/bioinformatics/btad056}

\bibitem[{Weininger(1988)}]{weininger1988smiles}
Weininger D (1988) Smiles, a chemical language and information system. 1. introduction to methodology and encoding rules. Journal of chemical information and computer sciences 28(1):31--36. \doi{10.1021/ci00057a005}

\bibitem[{Bento et~al.(2020)Bento, Hersey, F{\'e}lix, Landrum, Gaulton, Atkinson, Bellis, De~Veij, and Leach}]{bento2020open}
Bento AP, Hersey A, F{\'e}lix E et~al (2020) An open source chemical structure curation pipeline using rdkit. Journal of Cheminformatics 12:1--16. \doi{10.1186/s13321-020-00456-1}

\bibitem[{Yu and Koltun(2015)}]{yu2015multi}
Yu F, Koltun V (2015) Multi-scale context aggregation by dilated convolutions. arXiv preprint arXiv:151107122 \doi{10.48550/arXiv.1511.07122}

\bibitem[{Jin et~al.(2023)Jin, Wu, Chen, Pan, Wang, Xie, Quan, and Lyu}]{jin2023capla}
Jin Z, Wu T, Chen T et~al (2023) Capla: improved prediction of protein--ligand binding affinity by a deep learning approach based on a cross-attention mechanism. Bioinformatics 39(2):btad049. \doi{10.1093/bioinformatics/btad049}

\bibitem[{Zhang et~al.(2022)Zhang, Wang, and Chen}]{zhang2022predicting}
Zhang L, Wang CC, Chen X (2022) Predicting drug--target binding affinity through molecule representation block based on multi-head attention and skip connection. Briefings in Bioinformatics 23(6):bbac468. \doi{10.1093/bib/bbac468}

\bibitem[{Kingma and Ba(2014)}]{kingma2014adam}
Kingma DP, Ba J (2014) Adam: A method for stochastic optimization. arXiv preprint arXiv:14126980 \doi{10.48550/arXiv.1412.6980}

\bibitem[{Barbet and Huclier-Markai(2019)}]{barbet2019equilibrium}
Barbet J, Huclier-Markai S (2019) Equilibrium, affinity, dissociation constants, ic5o: Facts and fantasies. Pharmaceutical Statistics 18(5):513--525. \doi{10.1002/pst.1943}

\bibitem[{Lagunin et~al.(2018)Lagunin, Romanova, Zadorozhny, Kurilenko, Shilov, Pogodin, Ivanov, Filimonov, and Poroikov}]{lagunin2018comparison}
Lagunin AA, Romanova MA, Zadorozhny AD et~al (2018) Comparison of quantitative and qualitative (q) sar models created for the prediction of ki and ic50 values of antitarget inhibitors. Frontiers in Pharmacology 9:1136. \doi{10.3389/fphar.2018.01136}

\bibitem[{He et~al.(2023)He, Chen, and Chen}]{he2023nhgnn}
He H, Chen G, Chen CYC (2023) Nhgnn-dta: a node-adaptive hybrid graph neural network for interpretable drug--target binding affinity prediction. Bioinformatics 39(6):btad355. \doi{10.1093/bioinformatics/btad355}

\bibitem[{Li et~al.(2019)Li, Fu, and Zhang}]{li2019overview}
Li J, Fu A, Zhang L (2019) An overview of scoring functions used for protein--ligand interactions in molecular docking. Interdisciplinary Sciences: Computational Life Sciences 11:320--328. \doi{10.1007/s12539-019-00327-w}

\bibitem[{{\v{S}}krh{\'a}k et~al.(2024){\v{S}}krh{\'a}k, Novotn{\`y}, Feidakis, Kriv{\'a}k, and Hoksza}]{vskrhak2024cryptobench}
{\v{S}}krh{\'a}k V, Novotn{\`y} M, Feidakis CP et~al (2024) Cryptobench: Cryptic protein-ligand binding sites dataset and benchmark. Bioinformatics p btae745. \doi{10.1093/bioinformatics/btae745}

\bibitem[{Meller et~al.(2023)Meller, Ward, Borowsky, Kshirsagar, Lotthammer, Oviedo, Ferres, and Bowman}]{meller2023predicting}
Meller A, Ward M, Borowsky J et~al (2023) Predicting locations of cryptic pockets from single protein structures using the pocketminer graph neural network. Nature Communications 14(1):1177--1177. \doi{10.1038/s41467-023-36699-3}

\bibitem[{Corso et~al.(2022)Corso, St{\"a}rk, Jing, Barzilay, and Jaakkola}]{corso2022diffdock}
Corso G, St{\"a}rk H, Jing B et~al (2022) Diffdock: Diffusion steps, twists, and turns for molecular docking. arXiv preprint arXiv:221001776 \doi{10.48550/arXiv.2210.01776}

\bibitem[{Schneuing et~al.(2024)Schneuing, Harris, Du, Didi, Jamasb, Igashov, Du, Gomes, Blundell, Lio et~al.}]{schneuing2024structure}
Schneuing A, Harris C, Du Y et~al (2024) Structure-based drug design with equivariant diffusion models. Nature Computational Science 4(12):899--909. \doi{10.1038/s43588-024-00737-x}

\end{thebibliography}

\end{document}